\pdfoutput=1

\documentclass[11pt]{article}

\usepackage[final]{acl}

\usepackage{times}
\usepackage{latexsym}
\usepackage[T1]{fontenc}
\usepackage[utf8]{inputenc}
\usepackage{microtype}
\usepackage{inconsolata}
\usepackage{algorithm}
\usepackage{algpseudocode}

\usepackage{graphicx}
\usepackage{amsmath}
\usepackage{multirow}
\usepackage{xcolor}
\usepackage{array}
\usepackage{adjustbox}
\usepackage{dirtytalk}
\usepackage{hyperref}
\usepackage{rotating}
\usepackage{colortbl}
\usepackage{booktabs}
\usepackage{float}
\usepackage[section]{placeins}

\definecolor{Expert}{HTML}{FCE5CD}
\definecolor{Novice}{HTML}{C9DAF8}
\newcommand{\expert}[1]{\colorbox{Expert}{\hyperref[tab:participants]{E#1}}}
\newcommand{\novice}[1]{\colorbox{Novice}{\hyperref[tab:participants]{N#1}}}

\title{A Computational Method for Measuring ``Open Codes'' in Qualitative Analysis}

\author{
  John Chen \\
  University of Arizona, Tucson, AZ, USA \\
  \texttt{johnchen@arizona.edu} \\
  \textbf{Alexandros Lotsos, Sihan Cheng, Caiyi Wang, Lexie Zhao, Yanjia Zhang}, \\
  \textbf{Jessica Hullman, Bruce L. Sherin, Uri J. Wilensky, and Michael S. Horn} \\
  Northwestern University, Evanston, IL, USA \\
  \texttt{\{alexandroslotsos2026, sihancheng2026, GraceWang2025, xinyuezhao2020, fzhang\}@u.northwestern.edu} \\
  \texttt{\{jhullman, bsherin, uri, michael-horn\}@northwestern.edu} \\
}

\begin{document}
\maketitle
\begin{abstract}
Qualitative analysis is critical to understanding human datasets in many social science disciplines. A central method in this process is inductive coding, where researchers identify and interpret codes directly from the datasets themselves. Yet, this exploratory approach poses challenges for meeting methodological expectations (such as ``depth'' and ``variation''), especially as researchers increasingly adopt Generative AI (GAI) for support. Ground-truth-based metrics are insufficient because they contradict the exploratory nature of inductive coding, while manual evaluation can be labor-intensive. This paper presents a theory-informed computational method for measuring inductive coding results from humans and GAI. Our method first merges individual codebooks using an LLM-enriched algorithm. It measures each coder's contribution against the merged result using four novel metrics: Coverage, Overlap, Novelty, and Divergence. Through two experiments on a human-coded online conversation dataset, we 1) reveal the merging algorithm's impact on metrics; 2) validate the metrics' stability and robustness across multiple runs and different LLMs; and 3) showcase the metrics' ability to diagnose coding issues, such as excessive or irrelevant (hallucinated) codes. Our work provides a reliable pathway for ensuring methodological rigor in human-AI qualitative analysis. 
\end{abstract}

\section{Introduction}
Qualitative analysis is widely adopted across many social science disciplines. Most often, qualitative researchers apply descriptive labels (codes) in two ways: deductive coding, where codes are applied according to a preconceived coding scheme, and inductive coding (``open coding''), where codes are concepts derived from the raw data. 

While methodologies such as Grounded Theory (GT) \cite{corbin_chapter_2008, corbin_grounded_1990} and Thematic Analysis (TA) \cite{braun_using_2006, terry_thematic_2017} rely on the inductive approach to discover emergent patterns from human data, inductive coding is hampered by its subjective and time-consuming nature \cite{attride-stirling_thematic_2001, bowman_using_2023}. Since ``ground truth'' may not exist at this stage, truth-based evaluation methods (e.g., inter-coder reliability) mismatch inductive coding's inherent open-endedness \cite{mcdonald_reliability_2019, corbin_chapter_2008, terry_thematic_2017}. As many computational linguistics, machine learning, or qualitative research studies (e.g., \citeauthor{xiao_supporting_2023}) attempt to leverage Generative AI (GAI) for inductive coding tasks, a theory-informed and computationally operational evaluation method is urgently needed. 

This paper introduces a theory-informed computational method for systematically measuring open coding results from both human and machine coders. We propose an LLM-enhanced merging algorithm and four team-based evaluation metrics that do not rely on ground-truth assumptions. We conducted two empirical experiments to 1) reveal the merging algorithm’s impact on the metrics; 2) validate the metrics’ stability and robustness across multiple runs and different LLMs; and 3) showcase the metrics’ ability to diagnose coding issues, such as excessive or irrelevant (hallucinated) codes. The results illustrate the metrics' stability, robustness, and use cases, with caution notes regarding their limitations and potential risks.

\section{Related Work}
\subsection{The Nature and Challenges of Inductive Qualitative Coding}
During inductive coding, researchers identify concepts and themes directly from raw data \cite{strauss_basics_1998, rahman_advantages_2016}, aiming at discovering novel insights often without an existing theoretical framework \cite{corbin_chapter_2008, corbin_grounded_1990, strauss_basics_1998, terry_thematic_2017}. The process involves iterating through a corpus to identify meaningful segments and assign descriptive labels (codes) that emerge directly from the data. Researchers then often group their coded segments into labeled ``categories'' that are used for further analysis. This process resembles iterative clustering in machine learning contexts. Inductive coding is open-ended, subjective, and does not strive for a singular ``correct'' result \cite{terry_thematic_2017}. Rather, the process should capture as many aspects, patterns, or ``codable moments'' as possible \cite{terry_thematic_2017, corbin_grounded_1990, corbin_chapter_2008-1}.

Yet, inductive coding is inherently subjective, time-consuming, and prone to ambiguities, making methodological rigor difficult to achieve \cite{attride-stirling_thematic_2001, bowman_using_2023, braun_one_2021, bringer_maximizing_2004, tuckett_applying_2005, furniss_confessions_2011, saunders_saturation_2018}. Since the process aims to widely capture novel insights rather than enforcing consistency, deductive coding metrics (such as inter-rater reliability) become gravely inadequate due to their reliance on ``ground truth'' assumptions \cite{mcdonald_reliability_2019}. 

To address this mismatch, qualitative researchers are shifting towards team-based approaches, where the team constantly compares and contrasts codes from multiple individuals \cite{cascio_team-based_2019, thomas_general_2006}. Team-based approaches embrace different perspectives, resulting in more insights and mitigating individual biases \cite{corbin_grounded_1990, thomas_general_2006}. It makes researchers closer towards the elusive goals of inductive analysis, such as depth, variation, and theoretical saturation \cite{corbin_chapter_2008, adams_qualitative_2008, furniss_confessions_2011, saunders_saturation_2018}.

\subsection{Evaluating ML/GAI for Inductive Qualitative Coding}
ML/GAI approaches offer significant potential to support and enhance qualitative research by assisting in the coding process \cite{xiao_supporting_2023}. Existing computational approaches have primarily framed machine-assisted qualitative coding as either a classification-based task, which mimics human labels, or a generation-based task, which produces codes directly from data \cite{liew_optimizing_2014, gebreegziabher_patat_2023, rietz_cody_2021, xiao_supporting_2023, grootendorst2022bertopic, saravani2023automated, sievert2014ldavis, de_paoli_can_2023, sinha_role_2024}. 

Effectively leveraging ML/GAI's potential requires robust evaluation methods that account for the open-ended and exploratory characteristics of inductive coding. However, existing evaluation strategies remain largely misaligned with these goals:

\begin{enumerate}

    \item \textbf{``Ground truth''-based metrics} compare an input set of codes against an expert-labeled dataset \cite{parfenova2025text, zhao2024new, dai-etal-2023-llm}. While it provides quantifiable metrics such as precision and recall, its presupposition of a single correct answer directly contradicts qualitative research theories \cite{corbin_chapter_2008, terry_thematic_2017}. Essentially, this approach constrains inputs to a predefined scope, thereby limiting the discovery of novel insights \cite{liew_optimizing_2014, xiao_supporting_2023, parfenova2025text}.
    
    \item \textbf{Clustering and topic-coherence metrics} evaluate internal structure through measures such as compactness, separation, or word-level coherence \cite{rahimi2023contextualized}. Although they do not require ground truth, they prioritize internal homogeneity rather than conceptual breadth or interpretive variation. Inductive coding, by contrast, values the coverage of diverse and crosscutting ideas rather than tight cluster cohesion \cite{corbin_chapter_2008, terry_thematic_2017}. Moreover, these metrics treat each coder or model output in isolation, lacking a mechanism to assess how different coders align, diverge, or collectively represent the dataset \cite{thomas_general_2006}. Hence, their objectives are incompatible with the theoretical aims of inductive qualitative analysis.

    \item \textbf{Human-annotated evaluations} ask experts to assess the usefulness, explainability, or relevance of machine-generated codes through survey-based instruments \cite{de_paoli_can_2023, de_paoli_performing_2023, zambrano_ncoder_2023, spinoso2023qualitative}. While such methods provide valuable qualitative insights, they are costly, time-consuming, and difficult to scale for large datasets or iterative evaluation cycles, potentially overlooking systematic omissions when all coders fail to recognize a key concept \cite{parfenova2025text}.
\end{enumerate}


\section{Computational Metrics}
\label{sec:metrics}

Building on the team-based approach adopted by qualitative researchers \cite{thomas_general_2006}, our method aggregates multiple coders’ coding results (i.e., codebooks) into a shared analytical space and calculates four computational metrics. Guided by qualitative analysis theory and the limitations identified in existing evaluation methods, our metric design follows three core requirements: First, the evaluation must remain independent of any ground truth so that coders can be assessed even when no authoritative codebook exists. Second, it must capture both the breadth of ideas and the interpretive balance of each coder while discouraging over-fragmentation or redundant labeling. Third, it must be stable across different models and repeated runs while remaining sensitive to problematic cases such as hallucination or flooding. 

To address these goals, we introduce four complementary metrics: \textbf{Coverage}, \textbf{Overlap}, \textbf{Novelty}, and \textbf{Divergence}. Together, they evaluate how individual coders contribute to the collective interpretation of the data. Our method operates without assuming a ground truth or requiring human adjudication, though expert-coded results can serve as useful anchors when evaluating untested models or prompts. A reference implementation of the aggregation algorithm, full documentation, and metric calculation procedures are publicly available (\href{https://github.com/CIVITAS-John/LLM-Qualitative-Analysis}{open-sourced software package}, licensed under CC~BY--NC~4.0).

\subsection{Aggregating Coding Results: Code Spaces (CSP) and Aggregated Code Spaces (ACS)}
\begin{figure}[h]
    \centering
    \includegraphics[height=120pt]{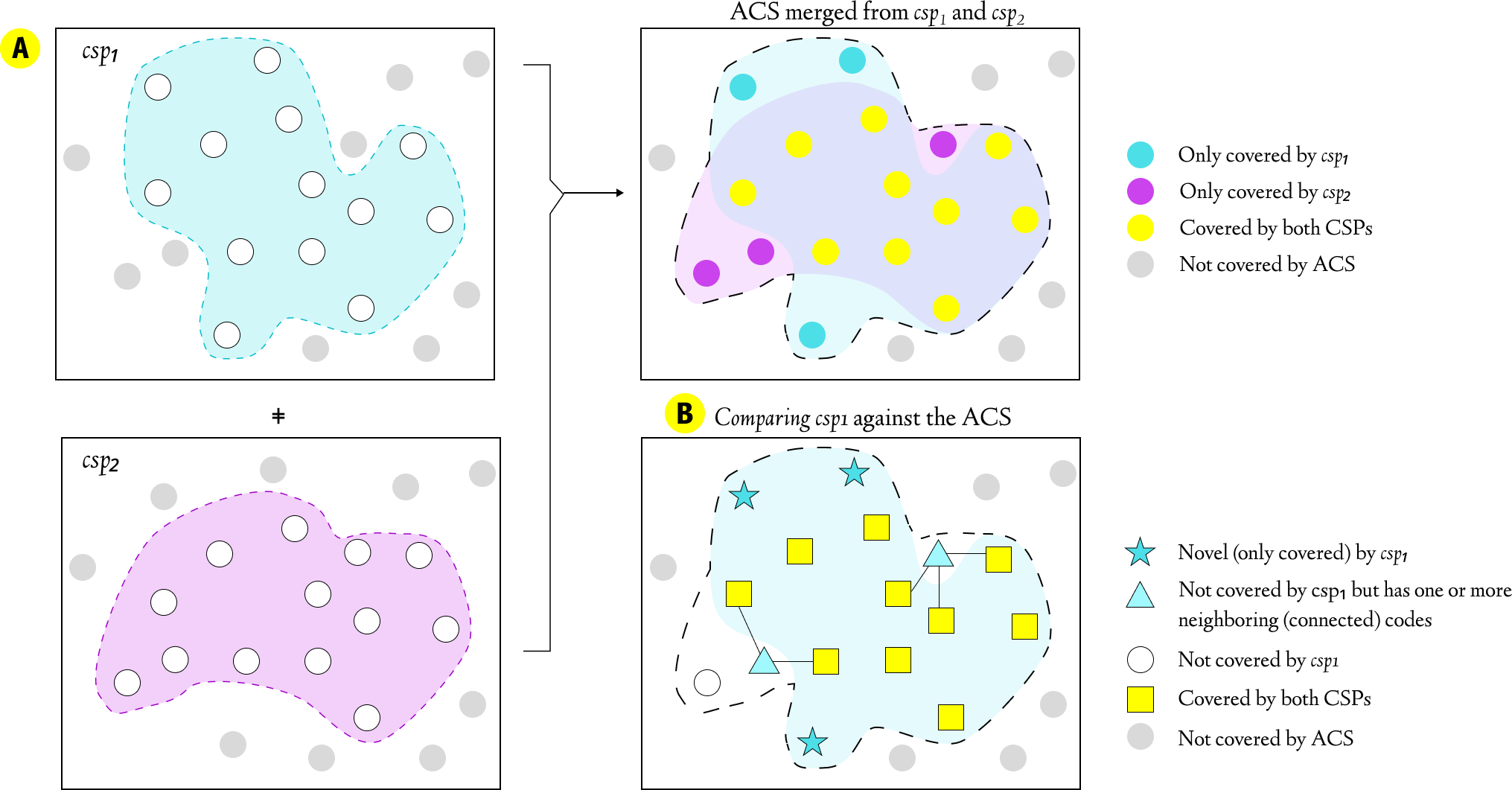}
    \caption{\textbf{A}: A conceptual illustration of an ACS merged from $csp_1$ and $csp_2$. \textbf{B}: Measuring $csp_1$ using the merged ACS as a reference.}
    \label{fig:acs-and-csp}
\end{figure}

The first step towards calculating our proposed metrics is to aggregate the codebooks produced by multiple individual coders into a single conceptual space that serves as an approximation of ``all possible interpretations'' of the data, as prescribed by qualitative analysis methods (Fig. \ref{fig:acs-and-csp}).

To do this, we first consider each coder's results and define their \textbf{Code Space} (CSP), to be the set of all codes they identified or interpreted from the dataset (Fig. \ref{fig:acs-and-csp}A). In turn, we can consider the union of all individual CSPs as the \textbf{Aggregate Code Space} (ACS) that encompasses the codes identified or interpreted from the dataset by all coders. However, simply considering the ACS to be this union does not take into account that real-world coders often use different codes to represent identical or very similar ideas (e.g., the codes ``User Feedback'' vs ``Feedback from User'').

To account for this, we propose a four-stage algorithm that utilizes semantic similarity and hierarchical clustering to iteratively consolidate an ACS from a set of individual CSPs. In addition to this, we also expand our definition of an ACS to include the fact that semantically similar codes are connected by links as $Neighbors$ (Fig. \ref{fig:acs-and-csp}B). Noting that each code in an individual CSP has a $label$ (the code itself) and \textit{may} have $examples$ (pieces of data the code was applied to), and a $definition$ (a description of the concept the code covers), the algorithm is thus defined as follows:

\begin{enumerate}
    \item Begin with the ACS simply being the union of all individual CSPs, only considering the $label$ of each code.
    \item For each code in the ACS, use hierarchical clustering to merge codes with semantically very similar $labels$. For each merging pair, the shorter label will be adopted.
    \item For each code in the ACS, use LLMs to generate a new $definition$ based on its $label$ and $examples$. Then, repeat the merging using both $label$ and $definition$, using LLMs to generate the resulting $label$ and $definition$ of the merged code.
    \item Finally, repeat stage 3 but iteratively and using our modified clustering algorithm (see below for more details).
\end{enumerate}

In stages 2 and 3, we apply a strict threshold for cosine distances between the text embeddings of each code. However, we found that a single threshold was often insufficient in separating different codes. In stage 4, we therefore adapt hierarchical clustering to apply two merge thresholds ($lower$ and $upper$, with $lower < upper$) on each dendrogram node, from which the $penalty$ coefficient is calculated. These penalty terms prevent merges that would conflate distinct concepts and ensure that smaller or sparser codebooks do not exert disproportionate influence in the merged structure.

\begin{algorithm}[H]
\caption{First Penalty on the Difference Between Examples}
\label{alg:distance}
\begin{algorithmic}[1]
\Require Candidate codes $A,B \in ACS$ with embeddings $c_a,c_b$ and example sets $E_A,E_B$
\State $e = \frac{|E_A \cap E_B|}{|E_A \cup E_B|}$ \Comment{$e$: example-overlap ratio (Jaccard)}
\State $dist_{a,b} = d(c_a,c_b) + penalty * e^2$ \Comment{$d$: cosine distance}
\end{algorithmic}
\end{algorithm}

\begin{algorithm}[H]
\caption{Second Penalty on Unique Examples of the Potential Merge}
\label{alg:merge}
\begin{algorithmic}[1]
\Require $dist$ = adjusted distance of the node
\Require $count$ unique examples of the node
\Require $count_{avg}, count_{max}$: average and maximum unique-example counts across candidate nodes
\State $o = max(\frac{count - count_{avg}}{count_{max} - count_{avg}}, 0)$
\If{$dist \le lower$}
    \State \Return YES
\ElsIf{$dist > upper$}
    \State \Return NO
\ElsIf{$dist + penalty * o^2 < upper$}
    \State \Return YES
\Else
    \State \Return NO \Comment{Considered as ``Neighbors''}
\EndIf
\end{algorithmic}
\end{algorithm}

\subsection{Four Computational Metrics}
\label{evaluation-metrics}
The four proposed metrics measure each coder's CSP against the ACS in four ways: Coverage, Overlap, Novelty, and Divergence (Fig. \ref{fig:acs-and-csp}B). In Algorithms \ref{alg:weighting}--\ref{alg:divergence}, $x$ denotes a coder, $c$ a code in $ACS$, $obs_{x,c}$ the observation score, and $score_c$ the aggregate code weight.

\begin{itemize}
    \item \textbf{Coverage}: How much conceptual space does a given CSP cover in the ACS? Both TA and GT strive for ``richness'' of codes, capturing depth and variation for further analysis \cite{corbin_chapter_2008, braun2013successful}. Each code is weighted by the number of coders who identified it, and each coder is weighted by their total number of codes to prevent inflation from over-coding. Coverage increases as a coder contributes more distinct codes that expand the shared conceptual space, reflecting broader interpretive reach. Excessively high values, particularly when accompanied by many codes, may indicate redundancy or flooding.
    
\begin{algorithm}[H]
\caption{Weighting Codes and Codebooks}
\label{alg:weighting}
\begin{algorithmic}[1]
\For{$x \in ACS$}
    \State $size_x = \max\{\#(code\in x), size_{median}\}$
    \State $weight_x = \frac{1}{\ln{(size_x)}}$
    \For{$code \in ACS$}
        \If{$c \in x$}
            \State $obs_{x,c}$ *= 1
        \Else
            \State $obs_{x,c}$ *= $\frac{\ln{(|neighbors \in x| + 1)}}{\ln{(|neighbors| + 1)}}$
        \EndIf
        \State $score_{c}$ += $obs_{x,c} * weight_x$
    \EndFor
\EndFor
\end{algorithmic}
\end{algorithm}

\begin{algorithm}[H]
\caption{Calculating Coverage}
\label{alg:coverage}
\begin{algorithmic}[1]
\For{$x \in ACS$}
    \State $coverage_x = \frac{\sum_{c\in acs} obs_{x,c} \times score_{c}}{\sum_{c\in acs} score_{c}}$
\EndFor
\end{algorithmic}
\end{algorithm}

    \item \textbf{Overlap}: How much does a given CSP overlap with others? Overlap increases as a coder’s codes coincide or neighbor those contributed by others, indicating stronger conceptual alignment and agreement. 
    Very high Overlap combined with low Novelty may suggest redundancy, whereas very low Overlap can indicate conceptual drift, inconsistency, or misunderstanding. The algorithm is almost the same as the one proposed for coverage, except each coder's impact on $score$ is removed from the ACS (see Divergence).
    
    \item \textbf{Novelty}: How many ``unique'' codes does a given CSP include (i.e., codes that no other coders identified)? Novelty increases when a coder introduces previously unseen ideas, capturing their contribution of new conceptual ground. Moderate Novelty indicates balanced and productive insight, while excessive Novelty with low Overlap may suggest idiosyncratic or hallucinatory coding. 
    In considering Novelty, we measure how much additional conceptual space a coder contributes beyond the team’s collective coverage.

\begin{algorithm}[H]
\caption{Calculating Novelty}
\label{alg:novelty}
\begin{algorithmic}[1]
\For{$x \in ACS$}
    \State $novelty_x = \frac{\sum_{c\in csp (novel = 1)} obs_{x, c} * score_c}{\sum_{c\in acs (novel = 1)} score_c}$ \Comment{$novel=1$: codes not identified by other coders}

\EndFor
\end{algorithmic}
\end{algorithm}
    
    \item \textbf{Divergence}: How far is a given CSP's code distribution from the ACS? Divergence increases as a coder’s focus or emphasis departs from the group’s shared distribution, signaling either distinctive analytical perspective or instability. 
    Lower Divergence indicates stronger alignment with the team baseline. We calculate each CSP's divergence as the separation from its probability distribution from that of other CSPs. We used the Jensen-Shannon Divergence (JSD) to tolerate potential zeros.
    
\begin{algorithm}[H]
\caption{Calculating Divergence}
\label{alg:divergence}
\begin{algorithmic}[1]
\For{$x \in ACS$}
    \State $B_c = score_c - obs_{x,c}*weight_x$\Comment Baseline - excluding the coder's contribution
    \State $divergence_x = \sqrt{JSD(B \parallel obs_{x,c})}$
\EndFor
\end{algorithmic}
\end{algorithm}
\end{itemize}

Because coding traditions vary, these metrics are best interpreted in combination rather than through fixed thresholds. In practice, productive coders tend to maintain moderate Coverage together with Overlap that reflects shared understanding and Novelty that contributes distinct yet relevant ideas. Exact ranges will vary by dataset and analytical context.

\section{Experimental Design}
To empirically validate our computational metrics and merging algorithm, we designed two experiments to answer three research questions (RQ1-RQ3):

\begin{enumerate}
    \item How does each stage of our merging algorithm affect our metrics?
    \item Given the probabilistic nature and different capabilities of LLMs, how robust or stable are our metrics?
    \item Can our metrics identify edge cases such as excessive codes or hallucinations?
\end{enumerate}

\subsection{Task and Dataset}
We reused the prompt, dataset, and human coding results from \citeauthor{chen2025processes}'s study, where researchers conducted manual evaluation. In each experiment, we applied our computational metrics to inductive qualitative coding results from four human coders (three PhD students, one undergraduate student, all in a U.S. higher education institution) and four machine coders (the same LLM with different prompts). Both experiments work on an online conversation dataset between Physics Lab (an online learning software)'s designers and teacher users. The conversation happened in public messaging groups, and the collection of such data has been approved by a university IRB. The dataset is attached as part of the Software package and has been properly anonymized. Our usage of the dataset is consistent with the original intention and IRB approval.

Due to human researchers' limited capacity, we focused on the first 127 messages. The same question was provided to human and machine coders: ``How did Physics Lab's online community emerge?'' 

Since most qualitative data are from human subjects and are subject to IRB protection, we intentionally chose open-source and locally available models for the experiments. We used Gemma3-27B (with temperature = 0.5) to generate new sets of machine codes. We used mxbai-embed-large to calculate semantic distances \cite{emb2024mxbai, li2023angle}. 

\subsection{Experiment 1: Ablation and Comparison Study}
Experiment 1 addresses RQ1 and RQ2 through an ablation study on the four stages of our merging algorithm \label{four-stages}. We chose four machine coders from \citeauthor{chen2025processes}'s comparison study: Chunk-Level (i.e., generate per ``chunk'' of messages); Chunk-Level, Structured; Item-Level (i.e., generate per message); Item-Level, Verb Phrases Only. 

\begin{enumerate}
    \item \textbf{Condition 1} corresponds to the first ``naive'' stage, where codes are merged solely by their labels. 
    \item \textbf{Condition 2} corresponds to the second stage, where codes are merged with a strict threshold (0.32) by their labels. 
    \item \textbf{Condition 3} corresponds to the third stage, where an LLM generates definitions based on each code's label and examples. Then, the codes are merged with a strict threshold (0.32) by labels and definitions. 
    \item \textbf{Condition 4} corresponds to the fourth stage, where the codes are iteratively merged with an upper threshold (0.55) and a lower threshold (0.32) by labels and definitions, until no more codes can be merged.
\end{enumerate}

For each condition and LLM used, we repeated 10 runs, recorded each human and machine coder's number of codes within the merged ACS, and calculated four computational metrics. In addition to individual coders, we also calculated metrics for the combination ``group'' of AI or human coders. Thresholds in each condition are chosen interactively through our example implementation. The strict threshold is chosen by ensuring that 10 code pairs with a semantic distance right below it have the same meanings. The upper threshold is chosen by ensuring that 10 code pairs right below it have at least similar meanings. 

In Stages 3 and 4, the study doubles as a comparison between different LLMs used in the process: Gemma3 27B (non-reasoning, small, open-source) \cite{gemma_2025}, Qwen QwQ 32B (reasoning, small, open-source) \cite{qwq32b}, GPT-4.1 (non-reasoning, large, proprietary), and Gemini-2.5-Pro (reasoning, large, proprietary). 

\subsection{Experiment 2: Measuring Edge Cases}
Experiment 2 addresses RQ3. Starting from the Item-Level coder, which performed the best (together with its Verb Phrases Only Variant) in human evaluation and computational metrics, we created three variants to simulate potential edge cases:

\begin{enumerate}
    \item \textbf{Flooding Coder} is explicitly instructed to generate an excessive number of codes per item.
    \item \textbf{Hallucinating Coder} has the same prompt but works with an irrelevant, AI-generated conversation.
    \item \textbf{Hallucinate + Flooding Coder} combines the two changes together.
\end{enumerate}

For each variant, we repeated 10 runs with the same human coders and machine coders, replacing results from the Item-Level coder with its variant. Since our preliminary results find little impact on LLM choice, we only used Gemma3 27B for this experiment. In total, both experiments cost \~8 million LLM tokens (5 read, 3 write), around \$20 for proprietary models.

\section{Empirical Study}
The following sections present our hypotheses and empirical results. We provide more details through Appendices \ref{appendices} and via the \href{https://github.com/CIVITAS-John/ACL-26-LLM-Qualitative}{reproduction repository}.

\subsection{RQ1: Measuring Each Stage's Impact on Our Merging Algorithm}
\label{rq1}
Experiment 1 first examines how each stage of our merging algorithm impacts each coder's number of merged codes and the resulting metrics.

\subsubsection{Hypothesis 1: Evaluation condition significantly affects the number of merged codes and computed metrics.}\label{h1}
We used ordinary least squares (OLS) regression to model the effect of algorithmic stages (Conditions 1-4) on the number of consolidated codes and each coder's four metrics.

Result: \textbf{Mostly Confirmed}. Each algorithm stage significantly reduced the total number of merged codes ($p < 0.001$). As shown in Table \ref{tab:h1-reg-coefs}, we observed significant shifts in computational metrics in Conditions 3 and 4, but not in 2.

\begin{table}[h]
\centering
\begin{adjustbox}{max width=\linewidth}
\begin{tabular}{|l|r|r|r|r|}
\hline
\textbf{Condition} & \textbf{Coverage} & \textbf{Overlap} & \textbf{Novelty} & \textbf{Divergence} \\ \hline
Condition 2 & \cellcolor{green!10}0.09\% & \cellcolor{red!10}-0.09\% & \cellcolor{green!10}0.05\% & \cellcolor{green!10}0.37\% \\
Condition 3 & \cellcolor{green!25}3.60\% & \cellcolor{green!25}5.45\% & \cellcolor{green!15}0.94\% & \cellcolor{red!25}-4.31\% \\
Condition 4 & \cellcolor{green!35}7.02\% & \cellcolor{green!35}7.86\% & \cellcolor{red!15}-1.64\% & \cellcolor{red!15}-1.91\% \\ \hline
\end{tabular}
\end{adjustbox}
\caption{OLS regression coefficients for evaluation metrics across Conditions 2 to 4 (relative to Condition 1). Conditions 1 and 2 are deterministic. For other values, p < 0.001.}
\label{tab:h1-reg-coefs}
\end{table}

\subsubsection{Hypothesis 2: Evaluation condition has minimal impact on the relative ranking of coder metrics.}
\label{h2}
We conducted a ranking stability analysis across conditions using one-way ANOVA with Tukey HSD post-hoc comparisons. 

Result: \textbf{Partially Confirmed}. While algorithmic stages shift the values of computational metrics, rankings remain relatively stable. For all metrics, rankings of top performers (\#1-5) stay the same. For other coders, rankings within the label-only (1, 2) and LLM-enriched conditions (3, 4) are highly similar. 

\subsection{RQ2: Evaluating the Robustness and Stability of Our Proposed Metrics}
\label{rq2}
Experiment 1 then evaluates whether our computational metrics remain robust across repeated runs and different LLMs in Conditions 3 and 4.

\subsubsection{Hypothesis 3: LLM used in the merging process significantly influences metrics and code counts.}
\label{h3}

We used OLS regression to model the effect of LLMs on the number of consolidated codes and each coder's four metrics, controlling for fixed effects between Conditions 3 and 4 and between individual coders.

\begin{figure}[H]
    \centering
    \includegraphics[width=\linewidth]{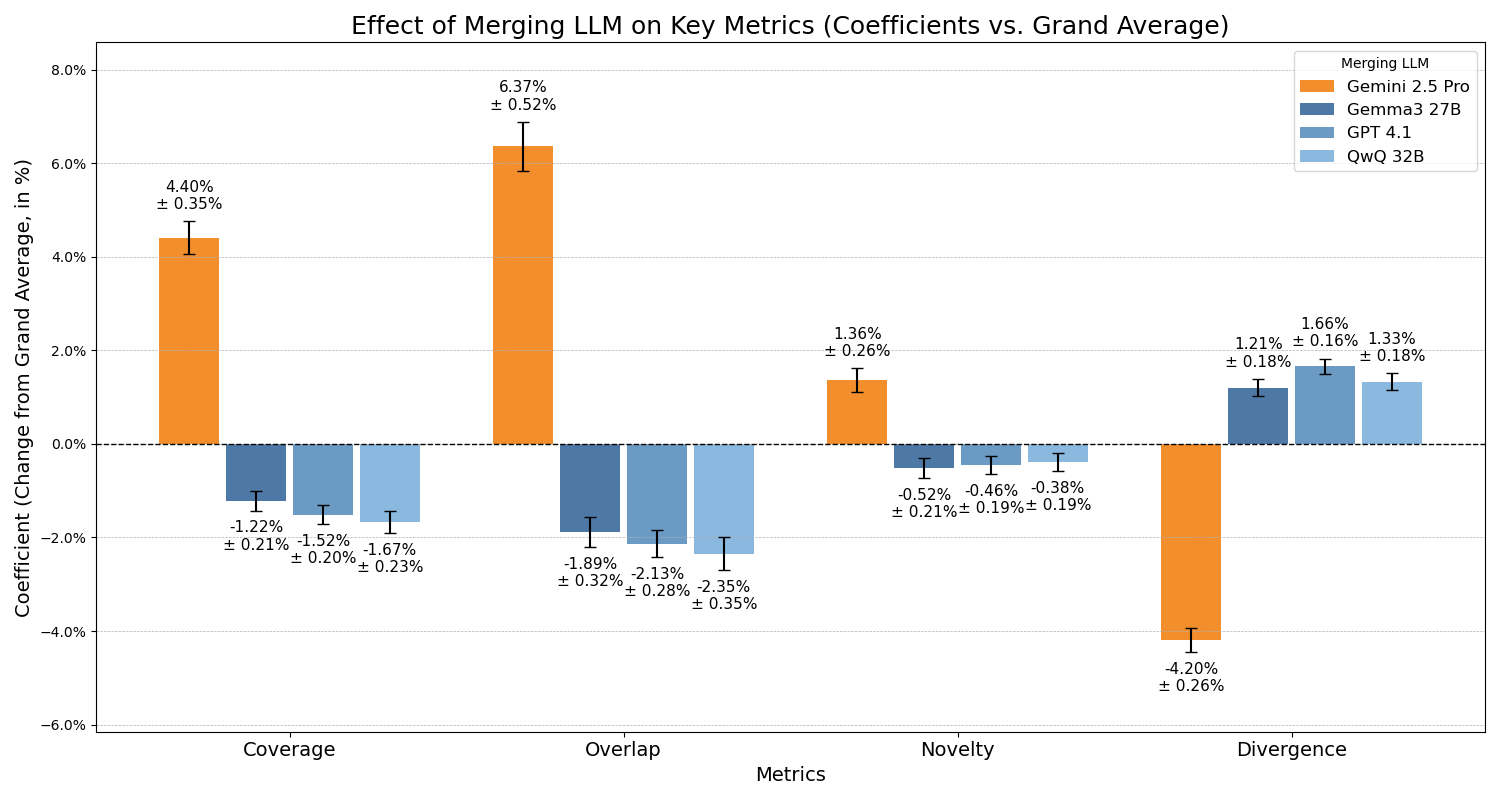}
    \caption{
        Effect of merging LLM on four evaluation metrics from the OLS model. Each bar represents the average coefficient difference from the grand mean across LLMs, along with 95\% confidence intervals.
    }
    \label{fig:merging-llm-effects}
\end{figure}

Result: \textbf{Partially Confirmed.}  
Across Conditions 3 and 4, three models (Gemma3 27B, Qwen QwQ 32B, and GPT 4.1) produce very similar metrics and numbers of merged codes. The only substantial deviation comes from \textit{Gemini-2.5-pro}, which produces fewer merged codes, higher coverage and overlap (approximately 4\% to 6\% increase), and lower divergence (4\% decrease) (Fig.~\ref{fig:merging-llm-effects}).

\subsubsection{Hypothesis 4: Metric outcomes are well-explained by condition, model, and coder identity.}
\label{h4}
From the same OLS model used by Hypothesis 3, we calculated adjusted~$R^2$ values to determine the extent to which the combination of condition, merging LLM, and coder explains variation in metric values.

Result: \textbf{Confirmed}. All adjusted~$R^2$ values exceed 0.91.

\subsubsection{Hypothesis 5: Repeated measurements under the same condition/model yield low coefficients of variation (CoV).}
\label{h5}
We calculated the coefficient of variation for each metric over 10 evaluation runs per Condition per merging LLM.

Result: \textbf{Confirmed}. CoV values remain below 0.1 in all cases. Divergence has the lowest variability around 0.01. Condition 4 shows slightly higher variance than Condition 3.

\subsubsection{Hypothesis 6: LLMs used in the merging process have little effect on the relative ranking of coders.}\label{h6}
We conducted one-way ANOVA with Tukey HSD post-hoc comparisons to examine whether coder rankings differed across LLMs.

Result: \textbf{Mostly Confirmed}. The top and bottom-ranked coders remained consistent across all four LLMs in both Conditions 3 and 4. Rankings for mid-performing coders fluctuate, primarily when their differences are not statistically significant.

\subsection{RQ3: Testing Our Proposed Metrics' Diagnostic Utility for Edge Cases}
\label{rq3}
\begin{figure}[H]
    \centering\includegraphics[width=\linewidth]{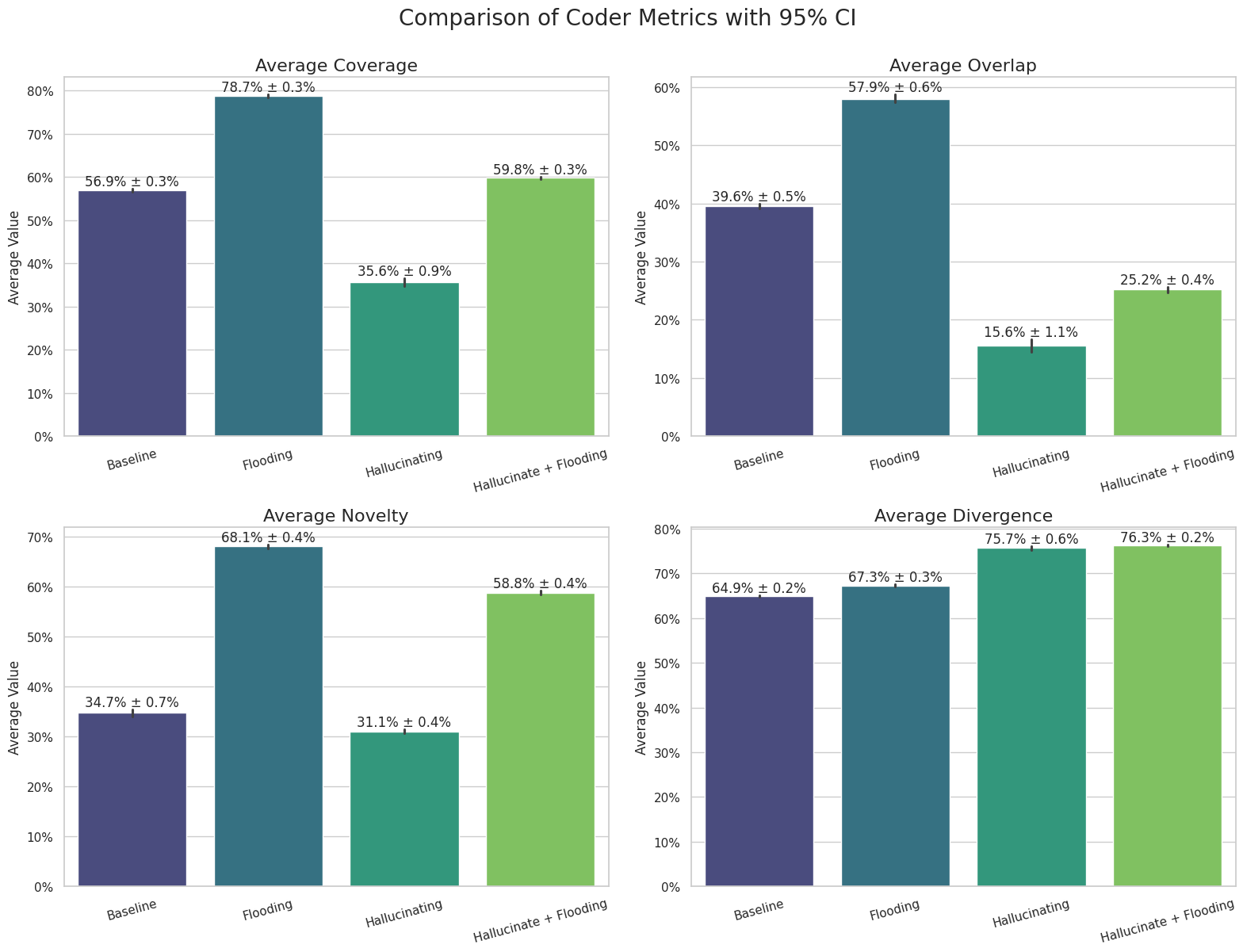}
    \caption{Mean coder metrics across Baseline, Flooding, Hallucinating, and Combined variants with 95\% confidence intervals.}
    \label{fig:edge-cases}
\end{figure}

Experiment 2 tests whether our computational metrics can detect abnormal inductive qualitative codes from machine coder variants designed to simulate edge cases: Flooding, Hallucinating, and Combined (Fig. \ref{fig:edge-cases}).

\subsubsection{Hypothesis 7: Excessive coding increases coverage, overlap, and novelty with diminishing returns, while divergence remains stable.}
Result: \textbf{Confirmed}. The mean metrics from the flooding coder have higher coverage (78.7\%), overlap (57.9\%), and novelty (68.1\%) than the Baseline coder. Novelty showed diminishing returns. Divergence remained stable (67.3\% vs. 64.9\%). A similar effect is observed between the Hallucinating and Combined coders. 

\subsubsection{Hypothesis 8: Coding on irrelevant data (i.e., hallucination) reduces coverage and overlap, while increasing divergence.}
Result: \textbf{Confirmed}. The mean metrics from the Hallucinating coder have reduced coverage (35.6\%) and overlap (15.6\%), while divergence increases sharply to 75.7\%. A stronger effect is observed between the Hallucinating and Combined coders. In particular, while the Combined coder produced 245\% more codes than the baseline (1,775 vs. 514), it has lower overlap (39.6\% vs. 25.2\%) and higher divergence (76.3\% vs. 64.9\%). 

\section{Discussions}
This paper introduces a theory-informed computational method for systematically evaluating the outputs of inductive coding. At the heart of our contribution are four metrics designed to capture the multifaceted nature of inductive coding:

\begin{itemize}
    \item \textbf{Coverage} measures the depth and variation of a coder's contribution against the Aggregated Code Space (ACS), reflecting the qualitative goal of achieving breadth and depth in analysis \cite{corbin_chapter_2008}.
    \item \textbf{Overlap} quantifies a coder's alignment with the conceptual consensus of the group, indicating how much their interpretations resonate with others.
    \item \textbf{Novelty} identifies the unique concepts a coder introduces, highlighting their potential value in bringing new perspectives to the analytical process.
    \item \textbf{Divergence} measures how much a coder's conceptual focus differs from others in the group, offering insight into their unique analytical lens.
\end{itemize}

Taken together, these metrics provide a holistic and nuanced assessment of a coder's performance that does not rely on ``ground truths'' that may not exist even with teams of human experts. Instead of pursuing a single, simplistic measure of agreement, our method embraces the subjectivity and exploratory spirit of inductive qualitative analysis \cite{corbin_grounded_1990, braun_thematic_2012}. It enables researchers to appreciate not only consensus but also the valuable variation that different coders bring to the table, providing a more complete picture of human-AI or human-human collaboration processes.

\subsection{The Necessity of the Iterative, LLM-Enriched Merging Algorithm}
RQ1 explored the impact of each stage of our merging algorithm. The findings from our ablation study (see \ref{rq1}) show that while each stage progressively reduces the number of codes in the Aggregated Code Space (ACS), the most significant shift in our metrics occurs between Stage 2 and Stage 3. This transition, which introduces LLM-generated definitions to the merging process, is far more impactful than simply relaxing semantic distance thresholds, as seen in the transitions between other stages.

This result confirms a central premise of our work: capturing true conceptual similarity in inductive coding requires moving beyond superficial linguistic parallels. The ``naive'' merging stages (Conditions 1 and 2), which rely solely on code labels, are insufficient. Therefore, the introduction of LLM-generated definitions centrally contributes to our proposed metrics. This stage infuses the semantic representation of each code with meaning derived from its underlying data points (``examples''), without forcing a direct comparison between those data points. Such insulation is vital for the inductive workflow, where coders may legitimately identify the same concept in different segments of the data \cite{corbin_grounded_1990}; comparing data points directly would be prone to error.

Hypotheses 1 and 2 also clarify the distinct roles of Stage 4 (iteratively merging with more relaxed thresholds), which produces a highly consolidated ACS but does not significantly impact the measuring outcome. Since we regenerate a label and definition for each merging pair, Stage 4 is computationally intensive. Therefore, its utility depends on the research goal. To produce a clean conceptual map for further interpretation, comparison, or synthesis (e.g., when identifying a coder's potential bias), the computational cost of Stage 4 is well justified. Conversely, when numerical metrics are the primary output (such as in large-scale model comparisons), the less computationally demanding Stage 3 may provide sufficiently stable and reliable outcomes for screening purposes.

\subsection{The Impact of LLM on Outcomes}
While the involvement of LLMs in the merging processes introduces intrinsic randomness, the findings from RQ2 confirm our method's reliability and robustness across multiple LLMs and repeated runs. As shown in Hypothesis 4, with very high adjusted $R^2$ values, our proposed metrics are overwhelmingly explained by coders' intrinsic merits. Repeated measurements under the same conditions yield low coefficients of variation (Hypothesis 5), showing stability across multiple LLM samplings.

Crucially, our metrics' stability extends to the choice of LLM. Tested on a diverse set of models, the choice of model has little effect on the relative ranking of coders (Hypothesis 6), where the top and bottom-ranked performers remained consistent across all LLMs. This implication is significant for the practical adoption by qualitative researchers, as they can confidently use smaller, open-source models (e.g., Gemma3 27B) to measure outcomes from human or machine coders. Such models can be easily deployed locally, providing better privacy protections and regulation compliance for handling often sensitive human subject data.

\subsection{Performance of Metrics in Edge Cases}
RQ3 evaluated the robustness and diagnostic utility of our metrics in simulated edge cases. The ``Flooding'' coder variant, explicitly prompted to produce an excessive number of codes, was expected to create significant redundancy. The metrics correctly capture that: both ``Flooding'' variants registered higher Coverage and Novelty, but with diminishing returns that mark the results' redundancy. 

Similarly, the ``Hallucinating'' coder variant, which worked from irrelevant data with the correct prompt, was expected to produce thematically similar codes but fail to capture critical details from the true data. The metrics capture the issue as well: both ``Hallucinating'' variants produced a predictable and dramatic drop in Coverage and Overlap, coupled with a sharp increase in Divergence.

In all three cases, our metrics behave predictably in response to those coding issues, suggesting that abnormal comparative values from our metrics can serve as a ``red flag'' for further evaluation. This diagnostic capability offers a layer of quality control that is essential for ensuring rigor in human-AI collaborative workflows.

\section{Conclusion}
This paper presents a reliable and robust computational method for the theory-informed, systematic evaluation of inductive coding. By moving beyond a reliance on a single ``ground truth,'' our approach provides a practical pathway for qualitative researchers to leverage AI in inductive analytical processes responsibly. By measuring and quantifying conceptual ideas like coverage, overlap, novelty, and divergence, we shift the evaluation focus of inductive coding from enforcing agreement to a more nuanced appreciation of the diverse contributions that each coder brings to the collaborative, exploratory process. As more and more qualitative researchers explore LLMs for inductive coding, we offer a timely contribution to ensure methodological rigor and facilitate more effective and transparent human-AI collaboration.

\section{Limitations}
While this study presents a robust computational method for evaluating inductive coding, it is important to acknowledge its limitations. From there, future research can help establish best practices and clear guidelines for qualitative researchers to adopt our metrics. Application and misuse risks are discussed separately in Appendix~\ref{appendix:potential-risks}.

\subsection{Limitation on the Dataset and Domain}
The validation of our metrics was conducted on a single dataset consisting of online conversations from a specific online community. While effective in this context, the method's performance and the metrics' utility need to be tested on more diverse forms of qualitative data, such as semi-structured interview transcripts or open-ended survey responses. The nuances of these different data types may present unique challenges not encountered in this study.

\subsection{Limitation on the Simulated Edge Cases}
The ``Flooding'' and ``Hallucinating'' coders were created through explicit instructions to simulate poor coding practices. Although our metrics successfully identified these simulated cases, further research is needed to validate their diagnostic capabilities in real-world scenarios. This includes testing the method on codes generated by novice human researchers, non-expert coders, or subtly flawed AI prompts, which may produce less extreme and harder-to-detect issues than our simulated variants. 

\subsection{Limitation on Missing Critical Codes}
Our framework evaluates coders relative to the Aggregated Code Space (ACS), which is constructed from the combined outputs of all coders. As a result, it cannot detect critical codes that are entirely absent from every individual code space. This blind spot is not unique to our method but inherent to all team-based or consensus-oriented qualitative evaluation approaches: if no coder identifies a particular concept, it will not appear in the collective representation. Ensuring coder diversity and using multiple human or model perspectives remain essential to mitigate this limitation.

\subsection{Limited Interpretations on Deviant LLM Behaviors}
Despite 3 out of 4 models behaving similarly in our experiment (Hypothesis 3), our findings observed deviated merging behaviors and behaviors with Gemini-2.5-Pro, which showed a tendency to merge codes more. Even when the ranking order is relatively stable, this model choice leads to significant changes in raw metric numbers. Further research should investigate whether and/or why the stronger reasoning model merges differently, and whether this deep merging may prove preferable or not to human researchers.

\section*{Acknowledgments}
We acknowledge the support and help from Dr. Matthew Berland, Dr. Lydia Cao, and Dr. James Spillane (and his students) during the research process. In addition, we acknowledge the support from the National Science Foundation under Grant Number 2303582. 

We also acknowledge the usage of Generative AI tools within and regarding the study. With constant human supervision and quality assurance, such tools are used to assist in: 

\begin{enumerate}
    \item Software development (e.g., copilot autocompletion; refactoring; generating boilerplate code or data cleanup code).
    \item Empirical data analysis (e.g., code for visualization or regression analysis based on explicit human instructions).
    \item Paper writing (e.g., proofreading, editing, LaTeX syntax support).
\end{enumerate}

%
\bibliography{bibliography}

\clearpage
\appendix
\onecolumn

\section{Potential Risks}
\label{appendix:potential-risks}
While our computational method is designed to mitigate risks in human-AI qualitative analysis, its improper application or the misinterpretation of its metrics can introduce other potential risks.

\subsection{Misinterpretation of Metrics}
Especially when coming from a non-qualitative research background, users may treat our computational metrics as objective, definitive measures of quality, contrary to the exploratory and subjective nature of inductive coding. For example, when substituting a baseline machine coder with its deviant variants, we interpreted the higher Divergence metrics as an indicator of potential hallucination. However, deviating from the consensus can also be valuable in other scenarios, such as a coder from a different intellectual tradition or lived background. Over-reliance on metric values can prevent junior researchers from developing the critical, interpretive skills that come from manual comparison, reflection, and building consensus with peers. Hence, our metrics should be used as diagnostic tools to prompt deeper qualitative inquiry, not as a substitute for it.

\subsection{Dependency on the Coding Team}
The Aggregated Code Space (ACS) is fundamentally a synthesis of the input codebook group. The quality and comprehensiveness of the evaluation are therefore dependent on the diversity and rigor of the coding team. If the entire human-AI team shares a particular bias or overlooks a critical theme, the ACS will reflect this omission. Consequently, researchers who rely solely on metric values risk automation bias, which diminishes critical engagement with the raw data and the coding process. We encourage researchers to continue recruiting a diverse human team and employing multiple LLMs for inductive coding. Moreover, they should also explore the interactive interface generated by our software package, which provides a network-based visualization of codes and metrics. We intend to further study the interface in an upcoming research project.

\subsection{Privacy and Data Security}
Our method involves processing data through Large Language Models and text embedding models. While our study prioritized the use of local, open-source models to protect sensitive data, researchers applying this method with proprietary, the usage of cloud-based APIs risk exposing confidential or personally identifiable information from their datasets. It is hence critical for researchers to maintain adherence to IRB protocols and relevant data privacy regulations.

\section{Appendices}
\label{appendices}
\subsection{Regression Results for Hypothesis 1}
\label{appendix:h1-regression}
This appendix provides complete regression tables supporting \textbf{Hypothesis 1} (Section~\ref{h1}), which tests whether the algorithmic condition significantly affects the number of consolidated codes and the quality of generated open codes. 

We report outputs from five ordinary least squares (OLS) models. \textbf{Regression 1} models the number of consolidated codes per transcript as a function of condition.  \textbf{Regression 2} consists of four separate models, each predicting one code-level metric: coverage, overlap, novelty, or divergence. Conditions are dummy-coded, and coder identity is sum-coded. All regressions use the design matrix described in Section~\ref{h1}, with coder identity sum-coded, condition dummy-coded (Condition~1 as the baseline), and heteroscedasticity-robust (HC3) standard errors.

\begin{table}[hbt!]
\small
\centering
\caption{Regression output for \textbf{Consolidated Code Count} (Hypothesis 1).}
\begin{tabular}{lrrrrr}
\textbf{Predictor} & \textbf{coef} & \textbf{std err} & \textbf{z} & \textbf{P$>|$z|} & \textbf{[0.025, 0.975]} \\
\hline
Intercept & 1509.00 & -- & -- & -- & [--, --] \\
Condition 2 & -54.00 & -- & -- & -- & [--, --] \\
Condition 3 & -110.88 & 3.01 & -36.90 & 0.000 & [-116.77, -104.99] \\
Condition 4 & -651.53 & 16.97 & -38.40 & 0.000 & [-684.78, -618.27] \\
\end{tabular}
\vspace{0.3em}
\end{table}
\begin{center}
\footnotesize\textit{Note: Standard errors unavailable for some parameters due to rank deficiency in constraints.}
\end{center}

\begin{table}[hbt!]
\small
\centering
\caption{Regression output for \textbf{Coverage \%} (Hypothesis 1).}
\begin{tabular}{lrrrrr}
\textbf{Predictor} & \textbf{coef} & \textbf{std err} & \textbf{z} & \textbf{P$>|$z|} & \textbf{[0.025, 0.975]} \\
\hline
Intercept & 28.89 & 0.80 & 34.20 & 0.000 & [27.20, 30.50] \\
Condition 2 & 0.09 & 1.20 & 0.08 & 0.936 & [-2.20, 2.40] \\
Condition 3 & 3.60 & 0.90 & 4.23 & 0.000 & [1.90, 5.30] \\
Condition 4 & 7.02 & 0.87 & 8.06 & 0.000 & [5.30, 8.70] \\
\textit{(Coder dummies omitted)} & & & & & \\
\end{tabular}
\end{table}

\begin{table}[hbt!]
\small
\centering
\caption{Regression output for \textbf{Overlap \%} (Hypothesis 1).}
\begin{tabular}{lrrrrr}
\textbf{Predictor} & \textbf{coef} & \textbf{std err} & \textbf{z} & \textbf{P$>|$z|} & \textbf{[0.025, 0.975]} \\
\hline
Intercept & 15.81 & 1.10 & 13.93 & 0.000 & [13.60, 18.00] \\
Condition 2 & -0.09 & 1.60 & -0.06 & 0.956 & [-3.30, 3.10] \\
Condition 3 & 5.45 & 1.10 & 4.75 & 0.000 & [3.20, 7.70] \\
Condition 4 & 7.86 & 1.20 & 6.69 & 0.000 & [5.60, 10.20] \\
\textit{(Coder dummies omitted)} & & & & & \\
\end{tabular}
\end{table}

\begin{table}[hbt!]
\small
\centering
\caption{Regression output for \textbf{Novelty \%} (Hypothesis 1).}
\begin{tabular}{lrrrrr}
\textbf{Predictor} & \textbf{coef} & \textbf{std err} & \textbf{z} & \textbf{P$>|$z|} & \textbf{[0.025, 0.975]} \\
\hline
Intercept & 23.95 & 0.20 & 97.69 & 0.000 & [23.50, 24.40] \\
Condition 2 & 0.05 & 0.30 & 0.15 & 0.879 & [-0.60, 0.70] \\
Condition 3 & 0.94 & 0.26 & 3.61 & 0.000 & [0.40, 1.40] \\
Condition 4 & -1.64 & 0.27 & -6.10 & 0.000 & [-2.20, -1.10] \\
\textit{(Coder dummies omitted)} & & & & & \\
\end{tabular}
\end{table}

\begin{table}[hbt!]
\small
\centering
\caption{Regression output for \textbf{Divergence \%} (Hypothesis 1).}
\begin{tabular}{lrrrrr}
\textbf{Predictor} & \textbf{coef} & \textbf{std err} & \textbf{z} & \textbf{P$>|$z|} & \textbf{[0.025, 0.975]} \\
\hline
Intercept & 72.35 & 0.60 & 125.55 & 0.000 & [71.20, 73.50] \\
Condition 2 & 0.37 & 0.80 & 0.48 & 0.633 & [-1.10, 1.90] \\
Condition 3 & -4.31 & 0.59 & -7.29 & 0.000 & [-5.50, -3.20] \\
Condition 4 & -1.91 & 0.60 & -3.20 & 0.001 & [-3.10, -0.70] \\
\textit{(Coder dummies omitted)} & & & & & \\
\end{tabular}
\end{table}

\FloatBarrier
\subsection{Coder Rankings for Hypothesis 2}
\label{appendix:h2-rankings}

This appendix supports \textbf{Hypothesis 2} (Section~\ref{h2}), which examines the relative ranking of coders across evaluation conditions. For each metric, we present the metric outcomes and rankings of 8 coders (plus two groups) per condition. \textbf{Rel} denotes the strength of evidence that the listed coder performs better than the next one in the ranking chain.

\textbf{Symbols:} $>>>$~ $p \leq 0.001$; $>>$~ $0.001 < p \leq 0.01$; $>$~ $0.01 < p \leq 0.05$; $\approx$ $p > 0.05$. These thresholds are derived from Tukey's HSD post-hoc comparisons following a one-way ANOVA on metric values, conducted separately for each condition. \textbf{Special note for Conditions 1 and 2}: those conditions do not involve probabilistic LLM-based merging and therefore are deterministic, denoted with $>>>$.

\definecolor{ai}{RGB}{220,235,247}
\definecolor{itemverb}{RGB}{255,242,204}
\definecolor{itemany}{RGB}{252,229,205}
\definecolor{human}{RGB}{222,239,217}
\definecolor{chunkstr}{RGB}{234,209,220}
\definecolor{chunkbar}{RGB}{255,217,217}

\newcommand{\colorrow}[2]{\cellcolor{#1}#2}

\begin{table}[h]
\small
\centering
\caption{Coder rankings by condition for \textbf{Coverage} (Hypothesis 2).\label{tab:h2-coverage}}
\resizebox{\textwidth}{!}{%
\begin{tabular}{llclclclc}
\toprule
\textbf{Condition 1} & \textbf{Rel} & \textbf{Condition 2} & \textbf{Rel} & \textbf{Condition 3} & \textbf{Rel} & \textbf{Condition 4} & \textbf{Rel} \\
\midrule
\colorrow{ai}{group: ai (84.80\%)} & $>>>$ & \colorrow{ai}{group: ai (84.85\%)} & $>>>$ & \colorrow{ai}{group: ai (87.66\%)} & $>>>$ & \colorrow{ai}{group: ai (87.29\%)} & $>>>$ \\
\colorrow{itemverb}{item-verb-ai (48.25\%)} & $>>>$ & \colorrow{itemany}{item-any-ai (48.46\%)} & $>>>$ & \colorrow{itemany}{item-any-ai (56.67\%)} & $>>>$ & \colorrow{itemany}{item-any-ai (60.48\%)} & $>>>$ \\
\colorrow{itemany}{item-any-ai (47.92\%)} & $>>>$ & \colorrow{itemverb}{item-verb-ai (48.24\%)} & $>>>$ & \colorrow{itemverb}{item-verb-ai (52.77\%)} & $>>>$ & \colorrow{itemverb}{item-verb-ai (57.11\%)} & $>>>$ \\
\colorrow{human}{group: human (40.58\%)} & $>>>$ & \colorrow{human}{group: human (40.22\%)} & $>>>$ & \colorrow{human}{group: human (43.93\%)} & $>>>$ & \colorrow{human}{group: human (47.10\%)} & $>>>$ \\
\colorrow{human}{human-c (15.15\%)} & $>>>$ & \colorrow{human}{human-c (15.16\%)} & $>>>$ & \colorrow{human}{human-c (18.36\%)} & $>>>$ & \colorrow{human}{human-c (23.83\%)} & $>>>$ \\
\colorrow{human}{human-b (14.10\%)} & $>>>$ & \colorrow{human}{human-b (13.90\%)} & $>>>$ & \colorrow{human}{human-b (16.11\%)} & $>>>$ & \colorrow{human}{human-b (20.36\%)} & $>>>$ \\
\colorrow{human}{human-a (10.65\%)} & $>>>$ & \colorrow{human}{human-a (10.96\%)} & $>>>$ & \colorrow{chunkstr}{chunk-structured-ai (13.68\%)} & $>$ & \colorrow{chunkstr}{chunk-structured-ai (17.32\%)} & $\approx$ \\
\colorrow{human}{human-d (10.18\%)} & $>>>$ & \colorrow{human}{human-d (10.34\%)} & $>>>$ & \colorrow{chunkbar}{chunk-barany-ai (12.64\%)} & $\approx$ & \colorrow{chunkbar}{chunk-barany-ai (16.60\%)} & $>$ \\
\colorrow{chunkbar}{chunk-barany-ai (8.90\%)} & $>>>$ & \colorrow{chunkbar}{chunk-barany-ai (8.89\%)} & $>>>$ & \colorrow{human}{human-d (12.43\%)} & $>>>$ & \colorrow{human}{human-d (15.25\%)} & $>$ \\
\colorrow{chunkstr}{chunk-structured-ai (8.38\%)} &       & \colorrow{chunkstr}{chunk-structured-ai (8.83\%)} &       & \colorrow{human}{human-a (10.71\%)} &       & \colorrow{human}{human-a (13.83\%)} &       \\
\bottomrule
\end{tabular}
}
\end{table}

\begin{table}[ht]
\small
\centering
\caption{Coder rankings by condition for \textbf{Overlap} (Hypothesis 2).}
\resizebox{\textwidth}{!}{%
\begin{tabular}{llclclclc}
\toprule
\textbf{Condition 1} & \textbf{Rel} & \textbf{Condition 2} & \textbf{Rel} & \textbf{Condition 3} & \textbf{Rel} & \textbf{Condition 4} & \textbf{Rel} \\
\midrule
\colorrow{ai}{group: ai (51.59\%)} & $>>>$ & \colorrow{ai}{group: ai (50.79\%)} & $>>>$ & \colorrow{ai}{group: ai (60.09\%)} & $>>>$ & \colorrow{ai}{group: ai (57.30\%)} & $>>>$ \\
\colorrow{itemany}{item-any-ai (29.22\%)} & $>>>$ & \colorrow{itemany}{item-any-ai (29.28\%)} & $>>>$ & \colorrow{itemany}{item-any-ai (41.64\%)} & $>>>$ & \colorrow{itemany}{item-any-ai (44.60\%)} & $>>>$ \\
\colorrow{itemverb}{item-verb-ai (28.09\%)} & $>>>$ & \colorrow{itemverb}{item-verb-ai (27.88\%)} & $>>>$ & \colorrow{itemverb}{item-verb-ai (36.49\%)} & $>>>$ & \colorrow{itemverb}{item-verb-ai (40.72\%)} & $>>>$ \\
\colorrow{human}{group: human (16.96\%)} & $>>>$ & \colorrow{human}{group: human (16.62\%)} & $>>>$ & \colorrow{human}{group: human (23.33\%)} & $>>>$ & \colorrow{human}{group: human (24.71\%)} & $>>>$ \\
\colorrow{human}{human-b (7.19\%)} & $>>>$ & \colorrow{human}{human-b (6.97\%)} & $>>>$ & \colorrow{human}{human-c (10.65\%)} & $\approx$ & \colorrow{human}{human-c (15.25\%)} & $>$ \\
\colorrow{human}{human-c (6.94\%)} & $>>>$ & \colorrow{human}{human-c (6.89\%)} & $>>>$ & \colorrow{human}{human-b (9.86\%)} & $\approx$ & \colorrow{human}{human-b (13.14\%)} & $\approx$ \\
\colorrow{human}{human-a (5.91\%)} & $>>>$ & \colorrow{human}{human-a (6.06\%)} & $>>>$ & \colorrow{chunkstr}{chunk-structured-ai (8.60\%)} & $\approx$ & \colorrow{chunkstr}{chunk-structured-ai (11.12\%)} & $\approx$ \\
\colorrow{human}{human-d (5.34\%)} & $>>>$ & \colorrow{human}{human-d (5.42\%)} & $>>>$ & \colorrow{human}{human-d (7.73\%)} & $\approx$ & \colorrow{chunkbar}{chunk-barany-ai (10.83\%)} & $\approx$ \\
\colorrow{chunkstr}{chunk-structured-ai (3.45\%)} & $>>>$ & \colorrow{chunkstr}{chunk-structured-ai (3.73\%)} & $>>>$ & \colorrow{chunkbar}{chunk-barany-ai (7.68\%)} & $\approx$ & \colorrow{human}{human-d (10.03\%)} & $\approx$ \\
\colorrow{chunkbar}{chunk-barany-ai (3.40\%)} &       & \colorrow{chunkbar}{chunk-barany-ai (3.56\%)} &       & \colorrow{human}{human-a (6.57\%)} &       & \colorrow{human}{human-a (8.97\%)} &       \\
\bottomrule
\end{tabular}
}
\end{table}

\begin{table}[ht]
\small
\centering
\caption{Coder rankings by condition for \textbf{Novelty} (Hypothesis 2).}
\resizebox{\textwidth}{!}{%
\begin{tabular}{llclclclc}
\toprule
\textbf{Condition 1} & \textbf{Rel} & \textbf{Condition 2} & \textbf{Rel} & \textbf{Condition 3} & \textbf{Rel} & \textbf{Condition 4} & \textbf{Rel} \\
\midrule
\colorrow{ai}{group: ai (80.50\%)} & $>>>$ & \colorrow{ai}{group: ai (80.44\%)} & $>>>$ & \colorrow{ai}{group: ai (81.03\%)} & $>>>$ & \colorrow{ai}{group: ai (77.74\%)} & $>>>$ \\
\colorrow{itemany}{item-any-ai (42.43\%)} & $>>>$ & \colorrow{itemany}{item-any-ai (42.20\%)} & $>>>$ & \colorrow{itemany}{item-any-ai (45.05\%)} & $>>>$ & \colorrow{itemany}{item-any-ai (36.59\%)} & $\approx$ \\
\colorrow{itemverb}{item-verb-ai (39.34\%)} & $>>>$ & \colorrow{itemverb}{item-verb-ai (39.69\%)} & $>>>$ & \colorrow{itemverb}{item-verb-ai (42.12\%)} & $>>>$ & \colorrow{itemverb}{item-verb-ai (35.80\%)} & $>>>$ \\
\colorrow{human}{group: human (30.17\%)} & $>>>$ & \colorrow{human}{group: human (30.25\%)} & $>>>$ & \colorrow{human}{group: human (30.84\%)} & $>>>$ & \colorrow{human}{group: human (27.12\%)} & $>>>$ \\
\colorrow{human}{human-c (11.65\%)} & $>>>$ & \colorrow{human}{human-c (11.49\%)} & $>>>$ & \colorrow{human}{human-c (12.84\%)} & $>>>$ & \colorrow{human}{human-c (11.78\%)} & $>>>$ \\
\colorrow{human}{human-b (9.43\%)} & $>>>$ & \colorrow{human}{human-b (9.46\%)} & $>>>$ & \colorrow{human}{human-b (9.77\%)} & $>>>$ & \colorrow{human}{human-b (8.47\%)} & $\approx$ \\
\colorrow{chunkstr}{chunk-structured-ai (7.44\%)} & $>>>$ & \colorrow{chunkstr}{chunk-structured-ai (7.54\%)} & $>>>$ & \colorrow{chunkstr}{chunk-structured-ai (7.79\%)} & $>$ & \colorrow{chunkstr}{chunk-structured-ai (8.00\%)} & $\approx$ \\
\colorrow{chunkbar}{chunk-barany-ai (7.20\%)} & $>>>$ & \colorrow{chunkbar}{chunk-barany-ai (7.21\%)} & $>>>$ & \colorrow{chunkbar}{chunk-barany-ai (7.21\%)} & $>>>$ & \colorrow{chunkbar}{chunk-barany-ai (7.36\%)} & $>>>$ \\
\colorrow{human}{human-d (6.14\%)} & $>>>$ & \colorrow{human}{human-d (6.25\%)} & $>>>$ & \colorrow{human}{human-d (6.44\%)} & $>$ & \colorrow{human}{human-d (5.37\%)} & $\approx$ \\
\colorrow{human}{human-a (5.25\%)} &       & \colorrow{human}{human-a (5.54\%)} &       & \colorrow{human}{human-a (5.82\%)} &       & \colorrow{human}{human-a (4.92\%)} &       \\
\bottomrule
\end{tabular}
}
\end{table}

\begin{table}[hbt!]
\small
\centering
\caption{Coder rankings by condition for \textbf{Divergence} (Hypothesis 2).}
\resizebox{\textwidth}{!}{%
\begin{tabular}{llclclclc}
\toprule
\textbf{Condition 1} & \textbf{Rel} & \textbf{Condition 2} & \textbf{Rel} & \textbf{Condition 3} & \textbf{Rel} & \textbf{Condition 4} & \textbf{Rel} \\
\midrule
\colorrow{chunkbar}{chunk-barany-ai (78.87\%)} & $>>>$ & \colorrow{chunkbar}{chunk-barany-ai (78.70\%)} & $>>>$ & \colorrow{human}{human-a (73.88\%)} & $>$ & \colorrow{human}{human-a (74.57\%)} & $\approx$ \\
\colorrow{chunkstr}{chunk-structured-ai (78.50\%)} & $>>>$ & \colorrow{chunkstr}{chunk-structured-ai (78.38\%)} & $>>>$ & \colorrow{chunkbar}{chunk-barany-ai (72.87\%)} & $\approx$ & \colorrow{human}{human-d (73.98\%)} & $\approx$ \\
\colorrow{human}{human-d (75.40\%)} & $>>>$ & \colorrow{human}{human-d (75.69\%)} & $>>>$ & \colorrow{human}{human-d (72.61\%)} & $>$ & \colorrow{chunkbar}{chunk-barany-ai (73.59\%)} & $\approx$ \\
\colorrow{human}{human-c (74.87\%)} & $>>>$ & \colorrow{human}{human-c (75.31\%)} & $>>>$ & \colorrow{chunkstr}{chunk-structured-ai (71.57\%)} & $\approx$ & \colorrow{chunkstr}{chunk-structured-ai (73.41\%)} & $>>$ \\
\colorrow{human}{human-a (74.72\%)} & $>>>$ & \colorrow{human}{human-a (75.10\%)} & $>>>$ & \colorrow{human}{human-c (71.24\%)} & $\approx$ & \colorrow{human}{human-b (72.37\%)} & $\approx$ \\
\colorrow{human}{human-b (74.09\%)} & $>>>$ & \colorrow{human}{human-b (74.67\%)} & $>>>$ & \colorrow{human}{human-b (70.89\%)} & $>>>$ & \colorrow{human}{human-c (71.86\%)} & $>>>$ \\
\colorrow{human}{group: human (69.45\%)} & $>>>$ & \colorrow{human}{group: human (69.94\%)} & $>>>$ & \colorrow{ai}{group: ai (65.24\%)} & $\approx$ & \colorrow{ai}{group: ai (70.05\%)} & $>>>$ \\
\colorrow{ai}{group: ai (68.84\%)} & $>>>$ & \colorrow{ai}{group: ai (69.45\%)} & $>>>$ & \colorrow{human}{group: human (64.77\%)} & $>>>$ & \colorrow{human}{group: human (68.09\%)} & $>>>$ \\
\colorrow{itemverb}{item-verb-ai (65.08\%)} & $>>>$ & \colorrow{itemverb}{item-verb-ai (65.56\%)} & $>>>$ & \colorrow{itemverb}{item-verb-ai (60.17\%)} & $>>>$ & \colorrow{itemverb}{item-verb-ai (64.07\%)} & $>>>$ \\
\colorrow{itemany}{item-any-ai (63.72\%)} &       & \colorrow{itemany}{item-any-ai (64.43\%)} &       & \colorrow{itemany}{item-any-ai (57.20\%)} &       & \colorrow{itemany}{item-any-ai (62.41\%)} &       \\
\bottomrule
\end{tabular}
}
\end{table}

\FloatBarrier
\subsection{Regression Results for Hypotheses 3 and 4}
\label{appendix:h3-h4-regression}

This appendix providesOLS  regression results supporting \textbf{Hypotheses 3} and \textbf{4} (Section~\ref{h3} and~\ref{h4}). Hypothesis~3 tests whether the LLM used in the merging process significantly influences each code-level metric and the number of consolidated codes. Hypothesis~4 evaluates how well condition, merging model, and coder identity together explain variation in outcomes. All results are based on the model:
\[
Y \sim \text{C(condition)} + \text{C(model, Sum)} + \text{C(coder, Sum)}
\]
where $Y$ represents each outcome of interest. We use dummy-coding for the condition (Condition 3 as the baseline), sum-coding for the model and coder identity (Average as the baseline), and HC3 heteroscedasticity-robust standard errors. Coefficients are interpreted as deviations from the baseline.

\begin{table}[hbt!]
\small
\centering
\caption{Regression output for \textbf{Consolidated Codes} (Hypotheses 3 and 4).}
\begin{tabular}{lrrrrr}
\textbf{Predictor} & \textbf{coef} & \textbf{std err} & \textbf{z} & \textbf{P$>|$z|} & \textbf{[0.025, 0.975]} \\
\hline
Intercept & 295.2700 & 3.015 & 97.94 & 0.000 & [289.361, 301.179] \\
Condition 4 & -78.1800 & 4.442 & -17.60 & 0.000 & [-86.886, -69.474] \\
Gemini-2.5-pro & -15.0150 & 4.609 & -3.26 & 0.001 & [-24.048, -5.982] \\
Gemma3-27b & 1.3700 & 3.642 & 0.38 & 0.707 & [-5.768, 8.508] \\
GPT-4.1 & 2.5150 & 3.607 & 0.70 & 0.486 & [-4.555, 9.585] \\
\textit{(Coder terms omitted)} & & & & & \\
\end{tabular}
\vspace{0.3em}
\begin{center}
\footnotesize{$R^2 = 0.951$ \quad Adjusted $R^2 = 0.950$}
\end{center}
\end{table}

\begin{table}[hbt!]
\small
\centering
\caption{Regression output for \textbf{Coverage \%} (Hypotheses 3 and 4).}
\begin{tabular}{lrrrrr}
\textbf{Predictor} & \textbf{coef} & \textbf{std err} & \textbf{z} & \textbf{P$>|$z|} & \textbf{[0.025, 0.975]} \\
\hline
Intercept & 0.3250 & 0.001 & 326.50 & 0.000 & [0.323, 0.327] \\
Condition 4 & 0.0342 & 0.002 & 22.59 & 0.000 & [0.031, 0.037] \\
Gemini-2.5-pro & 0.0440 & 0.002 & 24.48 & 0.000 & [0.041, 0.048] \\
Gemma3-27b & -0.0122 & 0.001 & -11.28 & 0.000 & [-0.014, -0.010] \\
GPT-4.1 & -0.0152 & 0.001 & -14.75 & 0.000 & [-0.017, -0.013] \\
\textit{(Coder terms omitted)} & & & & & \\
\end{tabular}
\vspace{0.3em}
\begin{center}
\footnotesize{$R^2 = 0.993$ \quad Adjusted $R^2 = 0.993$}
\end{center}
\end{table}

\begin{table}[hbt!]
\small
\centering
\caption{Regression output for \textbf{Overlap \%} (Hypotheses 3 and 4).}
\begin{tabular}{lrrrrr}
\textbf{Predictor} & \textbf{coef} & \textbf{std err} & \textbf{z} & \textbf{P$>|$z|} & \textbf{[0.025, 0.975]} \\
\hline
Intercept & 0.2126 & 0.001 & 147.81 & 0.000 & [0.210, 0.215] \\
Condition 4 & 0.0240 & 0.002 & 10.72 & 0.000 & [0.020, 0.028] \\
Gemini-2.5-pro & 0.0637 & 0.003 & 23.97 & 0.000 & [0.058, 0.069] \\
Gemma3-27b & -0.0189 & 0.002 & -11.55 & 0.000 & [-0.022, -0.016] \\
GPT-4.1 & -0.0213 & 0.001 & -14.74 & 0.000 & [-0.024, -0.018] \\
\textit{(Coder terms omitted)} & & & & & \\
\end{tabular}
\vspace{0.3em}
\begin{center}
\footnotesize{$R^2 = 0.970$ \quad Adjusted $R^2 = 0.969$}
\end{center}
\end{table}

\begin{table}[hbt!]
\small
\centering
\caption{Regression output for \textbf{Novelty \%} (Hypotheses 3 and 4).}
\begin{tabular}{lrrrrr}
\textbf{Predictor} & \textbf{coef} & \textbf{std err} & \textbf{z} & \textbf{P$>|$z|} & \textbf{[0.025, 0.975]} \\
\hline
Intercept & 0.2489 & 0.001 & 309.04 & 0.000 & [0.247, 0.251] \\
Condition 4 & -0.0258 & 0.001 & -20.21 & 0.000 & [-0.028, -0.023] \\
Gemini-2.5-pro & 0.0136 & 0.001 & 10.41 & 0.000 & [0.011, 0.016] \\
Gemma3-27b & -0.0052 & 0.001 & -4.75 & 0.000 & [-0.007, -0.003] \\
GPT-4.1 & -0.0046 & 0.001 & -4.59 & 0.000 & [-0.007, -0.003] \\
\textit{(Coder terms omitted)} & & & & & \\
\end{tabular}
\vspace{0.3em}
\begin{center}
\footnotesize{$R^2 = 0.994$ \quad Adjusted $R^2 = 0.994$}
\end{center}
\end{table}

\begin{table}[hbt!]
\small
\centering
\caption{Regression output for \textbf{Divergence \%} (Hypotheses 3 and 4).}
\begin{tabular}{lrrrrr}
\textbf{Predictor} & \textbf{coef} & \textbf{std err} & \textbf{z} & \textbf{P$>|$z|} & \textbf{[0.025, 0.975]} \\
\hline
Intercept & 0.6804 & 0.001 & 803.88 & 0.000 & [0.679, 0.682] \\
Condition 4 & 0.0240 & 0.001 & 20.54 & 0.000 & [0.022, 0.026] \\
Gemini-2.5-pro & -0.0420 & 0.001 & -31.99 & 0.000 & [-0.045, -0.039] \\
Gemma3-27b & 0.0121 & 0.001 & 13.04 & 0.000 & [0.010, 0.014] \\
GPT-4.1 & 0.0166 & 0.001 & 19.77 & 0.000 & [0.015, 0.018] \\
\textit{(Coder terms omitted)} & & & & & \\
\end{tabular}
\vspace{0.3em}
\begin{center}
\footnotesize{$R^2 = 0.920$ \quad Adjusted $R^2 = 0.919$}
\end{center}
\end{table}

\FloatBarrier
\subsection{Regression Results for Hypothesis 5}
\label{appendix:h5-regression}

This appendix provides regression outputs supporting \textbf{Hypothesis 5} (Section~\ref{h5}), which evaluates the stability of metric outputs by analyzing the coefficient of variation (CoV) across stochastic LLM runs in Conditions~3 and~4. All models use ordinary least squares (OLS) regression to predict the CoV for each metric, using condition, merging model, and coder identity as predictors. Condition is dummy-coded (Condition 3 as baseline), and both model and coder identity are sum-coded (average as baseline). Robust (HC3) standard errors are used.

\begin{table}[hbt!]
\small
\centering
\caption{Regression output for \textbf{CoV of Consolidated Code Count} (Hypothesis 5).}
\begin{tabular}{lrrrrr}
\textbf{Predictor} & \textbf{coef} & \textbf{std err} & \textbf{z} & \textbf{P$>|$z|} & \textbf{[0.025, 0.975]} \\
\hline
Intercept & 0.0040 & 0.001 & 5.93 & 0.000 & [0.003, 0.005] \\
Condition 4 & 0.0236 & 0.002 & 15.23 & 0.000 & [0.021, 0.027] \\
Gemini-2.5-pro & 0.0047 & 0.001 & 3.88 & 0.000 & [0.002, 0.007] \\
Gemma3-27b & -0.0023 & 0.002 & -1.49 & 0.136 & [-0.005, 0.001] \\
GPT-4.1 & 0.0021 & 0.001 & 1.55 & 0.120 & [-0.001, 0.005] \\
\textit{(Coder dummies omitted)} & & & & & \\
\multicolumn{6}{l}{$R^2 = 0.830$ \hspace{2em} Adjusted $R^2 = 0.797$} \\
\end{tabular}
\end{table}

\begin{table}[h]
\small
\centering
\caption{Regression output for \textbf{CoV of Coverage} (Hypothesis 5).}
\begin{tabular}{lrrrrr}
\textbf{Predictor} & \textbf{coef} & \textbf{std err} & \textbf{z} & \textbf{P$>|$z|} & \textbf{[0.025, 0.975]} \\
\hline
Intercept & 0.0272 & 0.003 & 9.36 & 0.000 & [0.021, 0.033] \\
Condition 4 & 0.0307 & 0.005 & 6.77 & 0.000 & [0.022, 0.040] \\
Gemini-2.5-pro & 0.0019 & 0.004 & 0.54 & 0.587 & [-0.005, 0.009] \\
Gemma3-27b & -0.0071 & 0.004 & -1.68 & 0.093 & [-0.015, 0.001] \\
GPT-4.1 & 0.0020 & 0.004 & 0.50 & 0.621 & [-0.006, 0.010] \\
\textit{(Coder dummies omitted)} & & & & & \\
\multicolumn{6}{l}{$R^2 = 0.737$ \hspace{2em} Adjusted $R^2 = 0.686$} \\
\end{tabular}
\end{table}

\begin{table}[hbt!]
\small
\centering
\caption{Regression output for \textbf{CoV of Overlap} (Hypothesis 5).}
\begin{tabular}{lrrrrr}
\textbf{Predictor} & \textbf{coef} & \textbf{std err} & \textbf{z} & \textbf{P$>|$z|} & \textbf{[0.025, 0.975]} \\
\hline
Intercept & 0.0440 & 0.004 & 11.34 & 0.000 & [0.036, 0.052] \\
Condition 4 & 0.0490 & 0.007 & 7.45 & 0.000 & [0.036, 0.062] \\
Gemini-2.5-pro & -0.0043 & 0.005 & -0.83 & 0.404 & [-0.014, 0.006] \\
Gemma3-27b & -0.0108 & 0.006 & -1.74 & 0.082 & [-0.023, 0.001] \\
GPT-4.1 & 0.0051 & 0.006 & 0.88 & 0.377 & [-0.006, 0.016] \\
\textit{(Coder dummies omitted)} & & & & & \\
\multicolumn{6}{l}{$R^2 = 0.698$ \hspace{2em} Adjusted $R^2 = 0.639$} \\
\end{tabular}
\end{table}

\begin{table}[hbt!]
\small
\centering
\caption{Regression output for \textbf{CoV of Novelty} (Hypothesis 5).}
\begin{tabular}{lrrrrr}
\textbf{Predictor} & \textbf{coef} & \textbf{std err} & \textbf{z} & \textbf{P$>|$z|} & \textbf{[0.025, 0.975]} \\
\hline
Intercept & 0.0359 & 0.004 & 8.09 & 0.000 & [0.027, 0.045] \\
Condition 4 & 0.0495 & 0.006 & 7.67 & 0.000 & [0.037, 0.062] \\
Gemini-2.5-pro & 0.0163 & 0.006 & 2.61 & 0.009 & [0.004, 0.029] \\
Gemma3-27b & -0.0130 & 0.006 & -2.18 & 0.029 & [-0.025, -0.001] \\
GPT-4.1 & -0.0049 & 0.005 & -1.01 & 0.311 & [-0.014, 0.005] \\
\textit{(Coder dummies omitted)} & & & & & \\
\multicolumn{6}{l}{$R^2 = 0.784$ \hspace{2em} Adjusted $R^2 = 0.741$} \\
\end{tabular}
\end{table}

\begin{table}[hbt!]
\small
\centering
\caption{Regression output for \textbf{CoV of Divergence} (Hypothesis 5).}
\begin{tabular}{lrrrrr}
\textbf{Predictor} & \textbf{coef} & \textbf{std err} & \textbf{z} & \textbf{P$>|$z|} & \textbf{[0.025, 0.975]} \\
\hline
Intercept & 0.0090 & 0.001 & 17.95 & 0.000 & [0.008, 0.010] \\
Condition 4 & 0.0038 & 0.001 & 4.80 & 0.000 & [0.002, 0.005] \\
Gemini-2.5-pro & 0.0042 & 0.001 & 5.61 & 0.000 & [0.003, 0.006] \\
Gemma3-27b & -0.0034 & 0.001 & -4.67 & 0.000 & [-0.005, -0.002] \\
GPT-4.1 & -0.0007 & 0.001 & -1.07 & 0.284 & [-0.002, 0.001] \\
\textit{(Coder dummies omitted)} & & & & & \\
\multicolumn{6}{l}{$R^2 = 0.597$ \hspace{2em} Adjusted $R^2 = 0.518$} \\
\end{tabular}
\end{table}

\FloatBarrier
\subsection{Coder Rankings for Hypothesis 6}
\label{appendix:h6-ranking}

This appendix presents full coder ranking tables supporting \textbf{Hypothesis~6} (Section~\ref{h6}), which investigates whether the choice of evaluation LLM affects the relative ranking of coders. For each metric, we compute average scores per coder and rank them separately under four evaluation models in \textbf{Condition~3} and \textbf{Condition~4}. \textbf{Rel} denotes the strength of evidence that the listed coder performs better than the next one in the ranking chain.

\textbf{Symbols:} $>>>$~ $p \leq 0.001$; $>>$~ $0.001 < p \leq 0.01$; $>$~ $0.01 < p \leq 0.05$; $\approx$ $p > 0.05$. These thresholds are derived from Tukey's HSD post-hoc comparisons following a one-way ANOVA on metric values, conducted separately for each condition.

\begin{table}[hbt!]
\small
\centering
\caption{Coder rankings for \textbf{Coverage}, Condition 3 (Hypothesis 6).}
\resizebox{\textwidth}{!}{%
\begin{tabular}{llclclclclc}
\toprule
\textbf{Condition 3} & \textbf{Rel} & \textbf{Gemini-2.5-pro} & \textbf{Rel} & \textbf{Gemma3-27B} & \textbf{Rel} & \textbf{GPT-4.1} & \textbf{Rel} & \textbf{Qwen-QwQ-32B} & \textbf{Rel} \\
\midrule
\colorrow{ai}{group: ai (87.66\%)} & $>>>$ & \colorrow{ai}{group: ai (90.23\%)} & $>>>$ & \colorrow{ai}{group: ai (86.50\%)} & $>>>$ & \colorrow{ai}{group: ai (86.64\%)} & $>>>$ & \colorrow{ai}{group: ai (87.25\%)} & $>>>$ \\
\colorrow{itemany}{item-any-ai (56.67\%)} & $>>>$ & \colorrow{itemany}{item-any-ai (60.27\%)} & $>$ & \colorrow{itemany}{item-any-ai (56.54\%)} & $>>>$ & \colorrow{itemany}{item-any-ai (54.61\%)} & $>>>$ & \colorrow{itemany}{item-any-ai (55.27\%)} & $>>>$ \\
\colorrow{itemverb}{item-verb-ai (52.77\%)} & $>>>$ & \colorrow{itemverb}{item-verb-ai (58.96\%)} & $>>>$ & \colorrow{itemverb}{item-verb-ai (51.35\%)} & $>>>$ & \colorrow{itemverb}{item-verb-ai (49.89\%)} & $>>>$ & \colorrow{itemverb}{item-verb-ai (50.87\%)} & $>>>$ \\
\colorrow{human}{group: human (43.93\%)} & $>>>$ & \colorrow{human}{group: human (48.08\%)} & $>>>$ & \colorrow{human}{group: human (42.39\%)} & $>>>$ & \colorrow{human}{group: human (42.34\%)} & $>>>$ & \colorrow{human}{group: human (42.90\%)} & $>>>$ \\
\colorrow{human}{human-c (18.36\%)} & $>>>$ & \colorrow{human}{human-c (22.85\%)} & $>>>$ & \colorrow{human}{human-c (16.47\%)} & $\approx$ & \colorrow{human}{human-c (16.79\%)} & $>>>$ & \colorrow{human}{human-c (17.32\%)} & $>>>$ \\
\colorrow{human}{human-b (16.11\%)} & $>>>$ & \colorrow{human}{human-b (18.11\%)} & $>>>$ & \colorrow{human}{human-b (16.14\%)} & $>>>$ & \colorrow{human}{human-b (14.80\%)} & $>>>$ & \colorrow{human}{human-b (15.37\%)} & $\approx$ \\
\colorrow{chunkstr}{chunk-structured-ai (13.68\%)} & $>$ & \colorrow{human}{human-d (13.64\%)} & $\approx$ & \colorrow{chunkstr}{chunk-structured-ai (14.29\%)} & $>>>$ & \colorrow{chunkstr}{chunk-structured-ai (12.71\%)} & $\approx$ & \colorrow{chunkstr}{chunk-structured-ai (14.69\%)} & $>>>$ \\
\colorrow{chunkbar}{chunk-barany-ai (12.64\%)} & $\approx$ & \colorrow{chunkstr}{chunk-structured-ai (13.02\%)} & $\approx$ & \colorrow{chunkbar}{chunk-barany-ai (12.83\%)} & $>>>$ & \colorrow{chunkbar}{chunk-barany-ai (11.83\%)} & $\approx$ & \colorrow{chunkbar}{chunk-barany-ai (13.15\%)} & $\approx$ \\
\colorrow{human}{human-d (12.43\%)} & $>>>$ & \colorrow{chunkbar}{chunk-barany-ai (12.74\%)} & $>$ & \colorrow{human}{human-d (12.02\%)} & $>>>$ & \colorrow{human}{human-d (11.62\%)} & $>>$ & \colorrow{human}{human-d (12.44\%)} & $>>>$ \\
\colorrow{human}{human-a (10.71\%)} &       & \colorrow{human}{human-a (11.39\%)} &       & \colorrow{human}{human-a (10.40\%)} &       & \colorrow{human}{human-a (10.52\%)} &       & \colorrow{human}{human-a (10.55\%)} &       \\
\bottomrule
\end{tabular}
}
\end{table}

\begin{table}[hbt!]
\small
\centering
\caption{Coder rankings for \textbf{Coverage}, Condition 4 (Hypothesis 6).}
\resizebox{\textwidth}{!}{%
\begin{tabular}{llclclclclc}
\toprule
\textbf{Condition 4} & \textbf{Rel} & \textbf{Gemini-2.5-pro} & \textbf{Rel} & \textbf{Gemma3-27B} & \textbf{Rel} & \textbf{GPT-4.1} & \textbf{Rel} & \textbf{Qwen-QwQ-32B} & \textbf{Rel} \\
\midrule
\colorrow{ai}{group: ai (87.29\%)} & $>>>$ & \colorrow{ai}{group: ai (91.26\%)} & $>>>$ & \colorrow{ai}{group: ai (85.28\%)} & $>>>$ & \colorrow{ai}{group: ai (87.15\%)} & $>>>$ & \colorrow{ai}{group: ai (85.45\%)} & $>>>$ \\
\colorrow{itemany}{item-any-ai (60.48\%)} & $>>>$ & \colorrow{itemany}{item-any-ai (69.10\%)} & $>>>$ & \colorrow{itemany}{item-any-ai (56.87\%)} & $>>>$ & \colorrow{itemany}{item-any-ai (59.19\%)} & $>>>$ & \colorrow{itemany}{item-any-ai (56.76\%)} & $>>>$ \\
\colorrow{itemverb}{item-verb-ai (57.11\%)} & $>>>$ & \colorrow{itemverb}{item-verb-ai (65.58\%)} & $>>>$ & \colorrow{itemverb}{item-verb-ai (54.42\%)} & $>>>$ & \colorrow{itemverb}{item-verb-ai (54.93\%)} & $>>>$ & \colorrow{itemverb}{item-verb-ai (53.49\%)} & $>>>$ \\
\colorrow{human}{group: human (47.10\%)} & $>>>$ & \colorrow{human}{group: human (55.52\%)} & $>>>$ & \colorrow{human}{group: human (44.90\%)} & $>>>$ & \colorrow{human}{group: human (44.16\%)} & $>>>$ & \colorrow{human}{group: human (43.83\%)} & $>>>$ \\
\colorrow{human}{human-c (23.83\%)} & $>>>$ & \colorrow{human}{human-c (32.22\%)} & $>>>$ & \colorrow{human}{human-c (20.99\%)} & $>>$ & \colorrow{human}{human-c (21.81\%)} & $>>>$ & \colorrow{human}{human-c (20.30\%)} & $>>>$ \\
\colorrow{human}{human-b (20.36\%)} & $>>>$ & \colorrow{human}{human-b (27.30\%)} & $>>>$ & \colorrow{human}{human-b (18.91\%)} & $>$ & \colorrow{human}{human-b (18.06\%)} & $>>>$ & \colorrow{human}{human-b (17.17\%)} & $\approx$ \\
\colorrow{chunkstr}{chunk-structured-ai (17.32\%)} & $\approx$ & \colorrow{chunkstr}{chunk-structured-ai (21.35\%)} & $\approx$ & \colorrow{chunkstr}{chunk-structured-ai (16.97\%)} & $\approx$ & \colorrow{chunkstr}{chunk-structured-ai (15.32\%)} & $\approx$ & \colorrow{chunkstr}{chunk-structured-ai (15.66\%)} & $\approx$ \\
\colorrow{chunkbar}{chunk-barany-ai (16.60\%)} & $>$ & \colorrow{chunkbar}{chunk-barany-ai (21.32\%)} & $\approx$ & \colorrow{chunkbar}{chunk-barany-ai (15.87\%)} & $>$ & \colorrow{chunkbar}{chunk-barany-ai (14.95\%)} & $\approx$ & \colorrow{chunkbar}{chunk-barany-ai (14.27\%)} & $\approx$ \\
\colorrow{human}{human-d (15.25\%)} & $>$ & \colorrow{human}{human-d (20.43\%)} & $\approx$ & \colorrow{human}{human-d (13.96\%)} & $\approx$ & \colorrow{human}{human-d (14.04\%)} & $\approx$ & \colorrow{human}{human-d (12.56\%)} & $\approx$ \\
\colorrow{human}{human-a (13.83\%)} &       & \colorrow{human}{human-a (18.83\%)} &       & \colorrow{human}{human-a (12.67\%)} &       & \colorrow{human}{human-a (12.43\%)} &       & \colorrow{human}{human-a (11.38\%)} &       \\
\bottomrule
\end{tabular}
}
\end{table}

\begin{table}[hbt!]
\small
\centering
\caption{Coder rankings for \textbf{Overlap}, Condition 3 (Hypothesis 6).}
\resizebox{\textwidth}{!}{%
\begin{tabular}{llclclclclc}
\toprule
\textbf{Condition 3} & \textbf{Rel} & \textbf{Gemini-2.5-pro} & \textbf{Rel} & \textbf{Gemma3-27B} & \textbf{Rel} & \textbf{GPT-4.1} & \textbf{Rel} & \textbf{Qwen-QwQ-32B} & \textbf{Rel} \\
\midrule
\colorrow{ai}{group: ai (60.09\%)} & $>>>$ & \colorrow{ai}{group: ai (70.62\%)} & $>>>$ & \colorrow{ai}{group: ai (55.60\%)} & $>>>$ & \colorrow{ai}{group: ai (56.08\%)} & $>>>$ & \colorrow{ai}{group: ai (58.08\%)} & $>>>$ \\
\colorrow{itemany}{item-any-ai (41.64\%)} & $>>>$ & \colorrow{itemany}{item-any-ai (47.70\%)} & $>>$ & \colorrow{itemany}{item-any-ai (41.21\%)} & $>>>$ & \colorrow{itemany}{item-any-ai (38.03\%)} & $>>>$ & \colorrow{itemany}{item-any-ai (39.60\%)} & $>>>$ \\
\colorrow{itemverb}{item-verb-ai (36.49\%)} & $>>>$ & \colorrow{itemverb}{item-verb-ai (45.72\%)} & $>>>$ & \colorrow{itemverb}{item-verb-ai (34.58\%)} & $>>>$ & \colorrow{itemverb}{item-verb-ai (31.84\%)} & $>>>$ & \colorrow{itemverb}{item-verb-ai (33.82\%)} & $>>>$ \\
\colorrow{human}{group: human (23.33\%)} & $>>>$ & \colorrow{human}{group: human (28.80\%)} & $>>>$ & \colorrow{human}{group: human (21.55\%)} & $>>>$ & \colorrow{human}{group: human (20.88\%)} & $>>>$ & \colorrow{human}{group: human (22.08\%)} & $>>>$ \\
\colorrow{human}{human-c (10.65\%)} & $\approx$ & \colorrow{human}{human-c (15.43\%)} & $>>>$ & \colorrow{human}{human-b (9.76\%)} & $\approx$ & \colorrow{human}{human-c (8.78\%)} & $\approx$ & \colorrow{human}{human-c (9.57\%)} & $\approx$ \\
\colorrow{human}{human-b (9.86\%)} & $\approx$ & \colorrow{human}{human-b (12.21\%)} & $>>>$ & \colorrow{chunkstr}{chunk-structured-ai (9.14\%)} & $\approx$ & \colorrow{human}{human-b (8.28\%)} & $\approx$ & \colorrow{chunkstr}{chunk-structured-ai (9.48\%)} & $\approx$ \\
\colorrow{chunkstr}{chunk-structured-ai (8.60\%)} & $\approx$ & \colorrow{human}{human-d (9.20\%)} & $\approx$ & \colorrow{human}{human-c (8.82\%)} & $>>$ & \colorrow{chunkstr}{chunk-structured-ai (7.44\%)} & $\approx$ & \colorrow{human}{human-b (9.18\%)} & $\approx$ \\
\colorrow{human}{human-d (7.73\%)} & $\approx$ & \colorrow{chunkstr}{chunk-structured-ai (8.37\%)} & $\approx$ & \colorrow{chunkbar}{chunk-barany-ai (7.85\%)} & $\approx$ & \colorrow{human}{human-d (6.77\%)} & $\approx$ & \colorrow{chunkbar}{chunk-barany-ai (8.06\%)} & $\approx$ \\
\colorrow{chunkbar}{chunk-barany-ai (7.68\%)} & $\approx$ & \colorrow{chunkbar}{chunk-barany-ai (8.14\%)} & $\approx$ & \colorrow{human}{human-d (7.34\%)} & $>>>$ & \colorrow{chunkbar}{chunk-barany-ai (6.68\%)} & $\approx$ & \colorrow{human}{human-d (7.62\%)} & $>$ \\
\colorrow{human}{human-a (6.57\%)} &       & \colorrow{human}{human-a (7.55\%)} &       & \colorrow{human}{human-a (6.19\%)} &       & \colorrow{human}{human-a (6.22\%)} &       & \colorrow{human}{human-a (6.30\%)} &       \\
\bottomrule
\end{tabular}
}
\end{table}

\begin{table}[hbt!]
\small
\centering
\caption{Coder rankings for \textbf{Overlap}, Condition 4 (Hypothesis 6).}
\resizebox{\textwidth}{!}{%
\begin{tabular}{llclclclclc}
\toprule
\textbf{Condition 4} & \textbf{Rel} & \textbf{Gemini-2.5-pro} & \textbf{Rel} & \textbf{Gemma3-27B} & \textbf{Rel} & \textbf{GPT-4.1} & \textbf{Rel} & \textbf{Qwen-QwQ-32B} & \textbf{Rel} \\
\midrule
\colorrow{ai}{group: ai (57.30\%)} & $>>>$ & \colorrow{ai}{group: ai (73.02\%)} & $>>>$ & \colorrow{ai}{group: ai (50.65\%)} & $>>>$ & \colorrow{ai}{group: ai (55.45\%)} & $>>>$ & \colorrow{ai}{group: ai (50.09\%)} & $>>>$ \\
\colorrow{itemany}{item-any-ai (44.60\%)} & $>>>$ & \colorrow{itemany}{item-any-ai (57.66\%)} & $>>$ & \colorrow{itemany}{item-any-ai (39.58\%)} & $>>>$ & \colorrow{itemany}{item-any-ai (42.19\%)} & $>>>$ & \colorrow{itemany}{item-any-ai (38.97\%)} & $>>>$ \\
\colorrow{itemverb}{item-verb-ai (40.72\%)} & $>>>$ & \colorrow{itemverb}{item-verb-ai (54.01\%)} & $>>>$ & \colorrow{itemverb}{item-verb-ai (36.58\%)} & $>>>$ & \colorrow{itemverb}{item-verb-ai (37.39\%)} & $>>>$ & \colorrow{itemverb}{item-verb-ai (34.92\%)} & $>>>$ \\
\colorrow{human}{group: human (24.71\%)} & $>>>$ & \colorrow{human}{group: human (34.62\%)} & $>>>$ & \colorrow{human}{group: human (21.65\%)} & $>>>$ & \colorrow{human}{group: human (21.66\%)} & $>>>$ & \colorrow{human}{group: human (20.90\%)} & $>>>$ \\
\colorrow{human}{human-c (15.25\%)} & $>$ & \colorrow{human}{human-c (24.31\%)} & $>>>$ & \colorrow{human}{human-c (12.09\%)} & $\approx$ & \colorrow{human}{human-c (13.18\%)} & $\approx$ & \colorrow{human}{human-c (11.40\%)} & $\approx$ \\
\colorrow{human}{human-b (13.14\%)} & $\approx$ & \colorrow{human}{human-b (20.23\%)} & $>>>$ & \colorrow{human}{human-b (11.52\%)} & $\approx$ & \colorrow{human}{human-b (10.89\%)} & $\approx$ & \colorrow{human}{human-b (9.92\%)} & $\approx$ \\
\colorrow{chunkstr}{chunk-structured-ai (11.12\%)} & $\approx$ & \colorrow{chunkbar}{chunk-barany-ai (15.39\%)} & $\approx$ & \colorrow{chunkstr}{chunk-structured-ai (10.89\%)} & $\approx$ & \colorrow{chunkbar}{chunk-barany-ai (9.10\%)} & $\approx$ & \colorrow{chunkstr}{chunk-structured-ai (9.72\%)} & $\approx$ \\
\colorrow{chunkbar}{chunk-barany-ai (10.83\%)} & $\approx$ & \colorrow{human}{human-d (15.11\%)} & $\approx$ & \colorrow{chunkbar}{chunk-barany-ai (10.15\%)} & $\approx$ & \colorrow{chunkstr}{chunk-structured-ai (9.01\%)} & $\approx$ & \colorrow{chunkbar}{chunk-barany-ai (8.68\%)} & $\approx$ \\
\colorrow{human}{human-d (10.03\%)} & $\approx$ & \colorrow{chunkstr}{chunk-structured-ai (14.88\%)} & $\approx$ & \colorrow{human}{human-d (8.65\%)} & $\approx$ & \colorrow{human}{human-d (8.98\%)} & $\approx$ & \colorrow{human}{human-d (7.37\%)} & $\approx$ \\
\colorrow{human}{human-a (8.97\%)} &       & \colorrow{human}{human-a (13.66\%)} &       & \colorrow{human}{human-a (7.78\%)} &       & \colorrow{human}{human-a (7.81\%)} &       & \colorrow{human}{human-a (6.62\%)} &       \\
\bottomrule
\end{tabular}
}
\end{table}

\begin{table}[hbt!]
\small
\centering
\caption{Coder rankings for \textbf{Novelty}, Condition 3 (Hypothesis 6).}
\resizebox{\textwidth}{!}{%
\begin{tabular}{llclclclclc}
\toprule
\textbf{Condition 3} & \textbf{Rel} & \textbf{Gemini-2.5-pro} & \textbf{Rel} & \textbf{Gemma3-27B} & \textbf{Rel} & \textbf{GPT-4.1} & \textbf{Rel} & \textbf{Qwen-QwQ-32B} & \textbf{Rel} \\
\midrule
\colorrow{ai}{group: ai (81.03\%)} & $>>>$ & \colorrow{ai}{group: ai (83.66\%)} & $>>>$ & \colorrow{ai}{group: ai (79.24\%)} & $>>>$ & \colorrow{ai}{group: ai (80.56\%)} & $>>>$ & \colorrow{ai}{group: ai (80.67\%)} & $>>>$ \\
\colorrow{itemany}{item-any-ai (45.05\%)} & $>>>$ & \colorrow{itemany}{item-any-ai (47.25\%)} & $>>>$ & \colorrow{itemany}{item-any-ai (44.34\%)} & $>>>$ & \colorrow{itemany}{item-any-ai (44.33\%)} & $>>>$ & \colorrow{itemany}{item-any-ai (44.26\%)} & $>>>$ \\
\colorrow{itemverb}{item-verb-ai (42.12\%)} & $>>>$ & \colorrow{itemverb}{item-verb-ai (44.94\%)} & $>>>$ & \colorrow{itemverb}{item-verb-ai (41.07\%)} & $>>>$ & \colorrow{itemverb}{item-verb-ai (40.69\%)} & $>>>$ & \colorrow{itemverb}{item-verb-ai (41.78\%)} & $>>>$ \\
\colorrow{human}{group: human (30.84\%)} & $>>>$ & \colorrow{human}{group: human (31.36\%)} & $>>>$ & \colorrow{human}{group: human (30.60\%)} & $>>>$ & \colorrow{human}{group: human (30.79\%)} & $>>>$ & \colorrow{human}{group: human (30.61\%)} & $>>>$ \\
\colorrow{human}{human-c (12.84\%)} & $>>>$ & \colorrow{human}{human-c (13.54\%)} & $>>>$ & \colorrow{human}{human-c (12.44\%)} & $>>>$ & \colorrow{human}{human-c (12.60\%)} & $>>>$ & \colorrow{human}{human-c (12.79\%)} & $>>>$ \\
\colorrow{human}{human-b (9.77\%)} & $>>>$ & \colorrow{human}{human-b (9.93\%)} & $>>$ & \colorrow{human}{human-b (9.76\%)} & $>>>$ & \colorrow{human}{human-b (9.97\%)} & $>>>$ & \colorrow{human}{human-b (9.44\%)} & $>>$ \\
\colorrow{chunkstr}{chunk-structured-ai (7.79\%)} & $>$ & \colorrow{chunkstr}{chunk-structured-ai (8.68\%)} & $\approx$ & \colorrow{chunkstr}{chunk-structured-ai (7.50\%)} & $>>$ & \colorrow{chunkstr}{chunk-structured-ai (7.65\%)} & $\approx$ & \colorrow{chunkstr}{chunk-structured-ai (7.34\%)} & $\approx$ \\
\colorrow{chunkbar}{chunk-barany-ai (7.21\%)} & $>>>$ & \colorrow{chunkbar}{chunk-barany-ai (7.90\%)} & $>$ & \colorrow{chunkbar}{chunk-barany-ai (6.86\%)} & $\approx$ & \colorrow{chunkbar}{chunk-barany-ai (7.08\%)} & $>>>$ & \colorrow{chunkbar}{chunk-barany-ai (6.99\%)} & $\approx$ \\
\colorrow{human}{human-d (6.44\%)} & $>$ & \colorrow{human}{human-d (6.77\%)} & $\approx$ & \colorrow{human}{human-d (6.44\%)} & $>$ & \colorrow{human}{human-d (6.09\%)} & $\approx$ & \colorrow{human}{human-d (6.46\%)} & $\approx$ \\
\colorrow{human}{human-a (5.82\%)} &       & \colorrow{human}{human-a (5.79\%)} &       & \colorrow{human}{human-a (5.94\%)} &       & \colorrow{human}{human-a (5.76\%)} &       & \colorrow{human}{human-a (5.80\%)} &       \\
\bottomrule
\end{tabular}
}
\end{table}

\begin{table}[h]
\small
\centering
\caption{Coder rankings for \textbf{Novelty}, Condition 4 (Hypothesis 6).}
\resizebox{\textwidth}{!}{%
\begin{tabular}{llclclclclc}
\toprule
\textbf{Condition 4} & \textbf{Rel} & \textbf{Gemini-2.5-pro} & \textbf{Rel} & \textbf{Gemma3-27B} & \textbf{Rel} & \textbf{GPT-4.1} & \textbf{Rel} & \textbf{Qwen-QwQ-32B} & \textbf{Rel} \\
\midrule
\colorrow{ai}{group: ai (77.74\%)} & $>>>$ & \colorrow{ai}{group: ai (80.88\%)} & $>>>$ & \colorrow{ai}{group: ai (75.22\%)} & $>>>$ & \colorrow{ai}{group: ai (78.34\%)} & $>>>$ & \colorrow{ai}{group: ai (76.50\%)} & $>>>$ \\
\colorrow{itemany}{item-any-ai (36.59\%)} & $\approx$ & \colorrow{itemany}{item-any-ai (39.56\%)} & $>>>$ & \colorrow{itemverb}{item-verb-ai (34.99\%)} & $\approx$ & \colorrow{itemany}{item-any-ai (36.49\%)} & $>$ & \colorrow{itemverb}{item-verb-ai (36.74\%)} & $\approx$ \\
\colorrow{itemverb}{item-verb-ai (35.80\%)} & $>>>$ & \colorrow{itemverb}{item-verb-ai (36.31\%)} & $>>>$ & \colorrow{itemany}{item-any-ai (34.74\%)} & $>>>$ & \colorrow{itemverb}{item-verb-ai (35.17\%)} & $>>>$ & \colorrow{itemany}{item-any-ai (35.58\%)} & $>>>$ \\
\colorrow{human}{group: human (27.12\%)} & $>>>$ & \colorrow{human}{group: human (27.41\%)} & $>>>$ & \colorrow{human}{group: human (28.28\%)} & $>>>$ & \colorrow{human}{group: human (25.06\%)} & $>>>$ & \colorrow{human}{group: human (27.73\%)} & $>>>$ \\
\colorrow{human}{human-c (11.78\%)} & $>>>$ & \colorrow{human}{human-c (11.97\%)} & $\approx$ & \colorrow{human}{human-c (11.84\%)} & $>>>$ & \colorrow{human}{human-c (11.14\%)} & $>>>$ & \colorrow{human}{human-c (12.16\%)} & $>>>$ \\
\colorrow{human}{human-b (8.47\%)} & $\approx$ & \colorrow{chunkstr}{chunk-structured-ai (11.29\%)} & $\approx$ & \colorrow{human}{human-b (8.70\%)} & $>>>$ & \colorrow{chunkstr}{chunk-structured-ai (8.26\%)} & $\approx$ & \colorrow{human}{human-b (8.03\%)} & $>>$ \\
\colorrow{chunkstr}{chunk-structured-ai (8.00\%)} & $\approx$ & \colorrow{chunkbar}{chunk-barany-ai (9.62\%)} & $\approx$ & \colorrow{chunkbar}{chunk-barany-ai (6.97\%)} & $\approx$ & \colorrow{human}{human-b (7.58\%)} & $\approx$ & \colorrow{chunkstr}{chunk-structured-ai (6.15\%)} & $\approx$ \\
\colorrow{chunkbar}{chunk-barany-ai (7.36\%)} & $>>>$ & \colorrow{human}{human-b (9.59\%)} & $>>>$ & \colorrow{chunkstr}{chunk-structured-ai (6.31\%)} & $\approx$ & \colorrow{chunkbar}{chunk-barany-ai (7.15\%)} & $>>>$ & \colorrow{chunkbar}{chunk-barany-ai (5.69\%)} & $\approx$ \\
\colorrow{human}{human-d (5.37\%)} & $\approx$ & \colorrow{human}{human-d (6.59\%)} & $\approx$ & \colorrow{human}{human-d (5.41\%)} & $\approx$ & \colorrow{human}{human-d (4.16\%)} & $\approx$ & \colorrow{human}{human-d (5.33\%)} & $\approx$ \\
\colorrow{human}{human-a (4.92\%)} &       & \colorrow{human}{human-a (6.24\%)} &       & \colorrow{human}{human-a (5.04\%)} &       & \colorrow{human}{human-a (4.09\%)} &       & \colorrow{human}{human-a (4.33\%)} &       \\
\bottomrule
\end{tabular}
}
\end{table}

\begin{table}[hbt!]
\small
\centering
\caption{Coder rankings for \textbf{Divergence}, Condition 3 (Hypothesis 6).}
\resizebox{\textwidth}{!}{%
\begin{tabular}{llclclclclc}
\toprule
\textbf{Condition 3} & \textbf{Rel} & \textbf{Gemini-2.5-pro} & \textbf{Rel} & \textbf{Gemma3-27B} & \textbf{Rel} & \textbf{GPT-4.1} & \textbf{Rel} & \textbf{Qwen-QwQ-32B} & \textbf{Rel} \\
\midrule
\colorrow{human}{human-a (73.88\%)} & $>$ & \colorrow{human}{human-a (71.91\%)} & $\approx$ & \colorrow{human}{human-a (74.56\%)} & $>>>$ & \colorrow{human}{human-a (74.62\%)} & $\approx$ & \colorrow{human}{human-a (74.44\%)} & $>>>$ \\
\colorrow{chunkbar}{chunk-barany-ai (72.87\%)} & $\approx$ & \colorrow{chunkbar}{chunk-barany-ai (71.78\%)} & $\approx$ & \colorrow{human}{human-c (73.38\%)} & $\approx$ & \colorrow{chunkbar}{chunk-barany-ai (74.52\%)} & $\approx$ & \colorrow{human}{human-d (72.83\%)} & $\approx$ \\
\colorrow{human}{human-d (72.61\%)} & $>$ & \colorrow{chunkstr}{chunk-structured-ai (71.29\%)} & $\approx$ & \colorrow{human}{human-d (73.05\%)} & $\approx$ & \colorrow{human}{human-d (74.13\%)} & $\approx$ & \colorrow{chunkbar}{chunk-barany-ai (72.40\%)} & $\approx$ \\
\colorrow{chunkstr}{chunk-structured-ai (71.57\%)} & $\approx$ & \colorrow{human}{human-d (70.45\%)} & $>>>$ & \colorrow{chunkbar}{chunk-barany-ai (72.76\%)} & $>>>$ & \colorrow{chunkstr}{chunk-structured-ai (73.53\%)} & $\approx$ & \colorrow{human}{human-c (72.38\%)} & $\approx$ \\
\colorrow{human}{human-c (71.24\%)} & $\approx$ & \colorrow{human}{human-b (67.73\%)} & $>>>$ & \colorrow{human}{human-b (71.07\%)} & $\approx$ & \colorrow{human}{human-c (73.43\%)} & $\approx$ & \colorrow{human}{human-b (71.78\%)} & $>>>$ \\
\colorrow{human}{human-b (70.89\%)} & $>>>$ & \colorrow{human}{human-c (65.79\%)} & $>>>$ & \colorrow{chunkstr}{chunk-structured-ai (71.01\%)} & $>>>$ & \colorrow{human}{human-b (72.96\%)} & $>>>$ & \colorrow{chunkstr}{chunk-structured-ai (70.44\%)} & $>>>$ \\
\colorrow{ai}{group: ai (65.24\%)} & $\approx$ & \colorrow{ai}{group: ai (60.48\%)} & $\approx$ & \colorrow{ai}{group: ai (66.89\%)} & $>$ & \colorrow{ai}{group: ai (67.14\%)} & $\approx$ & \colorrow{ai}{group: ai (66.46\%)} & $\approx$ \\
\colorrow{human}{group: human (64.77\%)} & $>>>$ & \colorrow{human}{group: human (59.93\%)} & $>>>$ & \colorrow{human}{group: human (66.34\%)} & $>>>$ & \colorrow{human}{group: human (66.88\%)} & $>>>$ & \colorrow{human}{group: human (65.93\%)} & $>>>$ \\
\colorrow{itemverb}{item-verb-ai (60.17\%)} & $>>>$ & \colorrow{itemverb}{item-verb-ai (53.78\%)} & $>>$ & \colorrow{itemverb}{item-verb-ai (61.38\%)} & $>>>$ & \colorrow{itemverb}{item-verb-ai (63.30\%)} & $>>>$ & \colorrow{itemverb}{item-verb-ai (62.21\%)} & $>>>$ \\
\colorrow{itemany}{item-any-ai (57.20\%)} &       & \colorrow{itemany}{item-any-ai (52.27\%)} &       & \colorrow{itemany}{item-any-ai (57.83\%)} &       & \colorrow{itemany}{item-any-ai (59.91\%)} &       & \colorrow{itemany}{item-any-ai (58.81\%)} &       \\
\bottomrule
\end{tabular}
}
\end{table}

\begin{table}[hbt!]
\small
\centering
\caption{Coder rankings for \textbf{Divergence}, Condition 4 (Hypothesis 6).}
\resizebox{\textwidth}{!}{%
\begin{tabular}{llclclclclc}
\toprule
\textbf{Condition 4} & \textbf{Rel} & \textbf{Gemini-2.5-pro} & \textbf{Rel} & \textbf{Gemma3-27B} & \textbf{Rel} & \textbf{GPT-4.1} & \textbf{Rel} & \textbf{Qwen-QwQ-32B} & \textbf{Rel} \\
\midrule
\colorrow{human}{human-a (74.57\%)} & $\approx$ & \colorrow{chunkstr}{chunk-structured-ai (70.94\%)} & $\approx$ & \colorrow{human}{human-a (75.75\%)} & $\approx$ & \colorrow{human}{human-a (75.38\%)} & $\approx$ & \colorrow{human}{human-a (76.45\%)} & $\approx$ \\
\colorrow{human}{human-d (73.98\%)} & $\approx$ & \colorrow{human}{human-a (70.68\%)} & $\approx$ & \colorrow{human}{human-d (75.28\%)} & $\approx$ & \colorrow{chunkstr}{chunk-structured-ai (75.00\%)} & $\approx$ & \colorrow{human}{human-d (75.98\%)} & $\approx$ \\
\colorrow{chunkbar}{chunk-barany-ai (73.59\%)} & $\approx$ & \colorrow{chunkbar}{chunk-barany-ai (70.36\%)} & $\approx$ & \colorrow{human}{human-c (74.41\%)} & $\approx$ & \colorrow{chunkbar}{chunk-barany-ai (74.76\%)} & $\approx$ & \colorrow{chunkbar}{chunk-barany-ai (75.01\%)} & $\approx$ \\
\colorrow{chunkstr}{chunk-structured-ai (73.41\%)} & $>>$ & \colorrow{human}{human-d (70.09\%)} & $>>>$ & \colorrow{chunkbar}{chunk-barany-ai (74.23\%)} & $\approx$ & \colorrow{human}{human-d (74.59\%)} & $\approx$ & \colorrow{human}{human-b (74.54\%)} & $\approx$ \\
\colorrow{human}{human-b (72.37\%)} & $\approx$ & \colorrow{human}{human-b (67.20\%)} & $>$ & \colorrow{human}{human-b (73.84\%)} & $\approx$ & \colorrow{human}{human-b (73.89\%)} & $\approx$ & \colorrow{human}{human-c (74.33\%)} & $\approx$ \\
\colorrow{human}{human-c (71.86\%)} & $>>>$ & \colorrow{human}{human-c (65.45\%)} & $\approx$ & \colorrow{chunkstr}{chunk-structured-ai (73.53\%)} & $>>$ & \colorrow{human}{human-c (73.23\%)} & $>>>$ & \colorrow{chunkstr}{chunk-structured-ai (74.18\%)} & $>>>$ \\
\colorrow{ai}{group: ai (70.05\%)} & $>>>$ & \colorrow{ai}{group: ai (65.00\%)} & $>>>$ & \colorrow{ai}{group: ai (72.02\%)} & $>>>$ & \colorrow{ai}{group: ai (71.33\%)} & $>>$ & \colorrow{ai}{group: ai (71.83\%)} & $>$ \\
\colorrow{human}{group: human (68.09\%)} & $>>>$ & \colorrow{human}{group: human (61.91\%)} & $>>>$ & \colorrow{human}{group: human (70.45\%)} & $>>>$ & \colorrow{human}{group: human (69.62\%)} & $>>>$ & \colorrow{human}{group: human (70.40\%)} & $>>>$ \\
\colorrow{itemverb}{item-verb-ai (64.07\%)} & $>>>$ & \colorrow{itemverb}{item-verb-ai (57.50\%)} & $\approx$ & \colorrow{itemverb}{item-verb-ai (66.28\%)} & $>>$ & \colorrow{itemverb}{item-verb-ai (65.87\%)} & $>>>$ & \colorrow{itemverb}{item-verb-ai (66.62\%)} & $>>>$ \\
\colorrow{itemany}{item-any-ai (62.41\%)} &       & \colorrow{itemany}{item-any-ai (56.40\%)} &       & \colorrow{itemany}{item-any-ai (64.94\%)} &       & \colorrow{itemany}{item-any-ai (63.87\%)} &       & \colorrow{itemany}{item-any-ai (64.45\%)} &       \\
\bottomrule
\end{tabular}
}
\end{table}

\end{document}